\documentclass[journal,twocolumn]{IEEEtran}
\usepackage{cite}

\usepackage{times}
\usepackage{epsfig}
\usepackage{graphicx}
\usepackage[tbtags]{amsmath}
\usepackage{amssymb}
\usepackage{float}
\usepackage[ruled,linesnumbered]{algorithm2e} 
\usepackage{algorithmicx}  
\usepackage{algpseudocode}  
\usepackage{enumerate}
\usepackage{authblk}
\usepackage{url}
\ifCLASSOPTIONcompsoc
\usepackage[caption=false,font=normalsize,labelfont=sf,textfont=sf]{subfig}
\else
\usepackage[caption=false,font=footnotesize]{subfig}
\fi
\usepackage{color}
\usepackage{ulem}

\usepackage{tikz}
\usetikzlibrary{decorations.pathreplacing}
\usetikzlibrary{positioning,arrows.meta,quotes}
\usetikzlibrary{shapes,snakes}
\usetikzlibrary{bayesnet}
\tikzset{>=latex}
\tikzstyle{plate caption} = [caption, node distance=0, inner sep=0pt, below left=5pt and 0pt of #1.south]

\usepackage[colorlinks,urlcolor=blue,linkcolor=black,anchorcolor=black,citecolor=black,breaklinks=true]{hyperref}
\usepackage{pdfpages}

\begin{document}
%
\title{Spatiotemporal Learning of Multivehicle Interaction Patterns in Lane-Change Scenarios 
}
%
%

\author{Chengyuan~Zhang,~\IEEEmembership{Student Member,~IEEE,}
        Jiacheng~Zhu,
        Wenshuo~Wang,~\IEEEmembership{Member,~IEEE},
        Junqiang Xi

\thanks{(Corresponding authors: Wenshuo Wang)}
\thanks{C. Zhang is with the Department of Civil Engineering, McGill University, Montreal, QC H3A 0C3, Canada. (e-mail: enzozcy@gmail.com).}
\thanks{J. Zhu is with the Department of Mechanical Engineering, Carnegie Mellon University, Pittsburgh, PA, USA, 15213. (e-mail: jzhu4@andrew.cmu.edu).}
\thanks{W. Wang is with the California Partners for Advanced Transportation Technology (PATH), UC Berkeley, USA. (e-mail: wwsbit@gmail.com).}
\thanks{J. Xi is with the Department of Mechanical Engineering, Beijing Institute of Technology, Beijing, China. (e-mail: xijunqiang@bit.edu.cn).}

}
\maketitle

\begin{abstract}
Interpretation of common-yet-challenging interaction scenarios can benefit well-founded decisions for autonomous vehicles. Previous research achieved this using their prior knowledge of specific scenarios with predefined models, limiting their adaptive capabilities. This paper describes a Bayesian nonparametric approach that leverages continuous (i.e., Gaussian processes) and discrete (i.e., Dirichlet processes) stochastic processes to reveal underlying interaction patterns of the ego vehicle with other nearby vehicles. Our model relaxes dependency on the number of surrounding vehicles by developing an acceleration-sensitive velocity field based on Gaussian processes. The experiment results demonstrate that the velocity field can represent the \textit{spatial} interactions between the ego vehicle and its surroundings.
Then, a discrete Bayesian nonparametric model, integrating Dirichlet processes and hidden Markov models, is developed to learn the interaction patterns over the \textit{temporal} space by segmenting and clustering the sequential interaction data into interpretable granular patterns automatically. We then evaluate our approach in the highway lane-change scenarios using the highD dataset collected from real-world settings. Results demonstrate that our proposed Bayesian nonparametric approach provides an insight into the complicated lane-change interactions of the ego vehicle with multiple surrounding traffic participants based on the interpretable interaction patterns and their transition properties in temporal relationships. Our proposed approach sheds light on efficiently analyzing other kinds of multi-agent interactions, such as vehicle-pedestrian interactions. View the demos via: \texttt{\url{https://youtu.be/z_vf9UHtdAM}}.
\end{abstract}


%


\section{Introduction}


\IEEEPARstart{C}{hanging} lanes in complex and diverse scenarios has become stubborn bottlenecks in the safe deployment of autonomous vehicles due to the behavioral uncertainty of nearby human-related agents \cite{galceran2017multipolicy,zhao2019influence}. According to their perception and experiences, human drivers can react to fast-varying surrounding traffics adaptively but could not exhaustively enumerate explicit rules for all scenarios from collected raw data solely as the flood of data can overwhelm human insight and analysis \cite{appenzeller2017scientist}. This is one of the core reasons why the specific models designed according to prior knowledge for autonomous vehicles can not deal with all real-world scenarios. Some research tailored decision-making policies for  predefined circumstances under specific settings, for example, only considering the  closest vehicles on the current and adjacent lanes, making behavior modeling and controller design feasible. For instance,  Zhang {\it et al}.\cite{zhang2018virtual} adopted distances to the nearest vehicles to reflect the local spatial relationship in traffic simulation, Leonhardt {\it et al}.\cite{leonhardt2017feature} evaluated a set of features about those closest vehicles on both the current and nearby lanes to encode the driving environment and status. However, they are far away from the real-world cases in which human drivers usually make decisions according to their comprehensive evaluation of surroundings in a specific spatiotemporal range\cite{schwarting2019social}. 

Researchers have developed several advanced machine learning techniques to model the interaction between the ego vehicle and its surroundings. However, most of these techniques are only suitable for limited scenarios with  a fixed number of nearby vehicles or with only the nearest vehicles into consideration. For example, the Bayesian networks in  \cite{kasper2012object,li2018estimating} required well-structured road conditions and only considered the nearest well-observed vehicles. Similar limitations also exist in the game theory of modeling multi-agent interactions\cite{schwarting2019social}. On the contrary, in the real-world, human drivers would comprehensively consider the entire scenario information within a region of interest (ROI) to make a proper decision; rather than only using the nearest vehicles' information\cite{kamal2015efficient}. However, the number of vehicles in the ROI is ordinarily changing over time, bringing significant challenges in modeling the interaction behaviors.

On the other hand, interaction pattern analysis of lane-change scenarios can help researchers design associated decision-making policies and facilitate to reveal the mechanism of changing lanes. However, there is still no unified specific rule of doing pattern analysis for lane-change interaction behavior. Some researchers prefer the grid-based method over the \textit{spatial} space in which a preset occupancy grid describes the surrounding space around the ego vehicle\cite{do2017human,li2018estimating}, but it fails to characterize the behavioral sequences over time. Some other works use the segment-based method over the \textit{temporal} space in which several manually predefined stages over time describe the entire lane-change behavior \cite{galceran2017multipolicy}. However, it only focused on the ego vehicle's maneuvers/trajectories while ignoring the interaction with other vehicles. Besides, both of the above ways are subjectively defined by researchers based on their prior knowledge and tailored applications. Therefore, given a bunch of lane-change scenarios in the real world, it is not yet entirely clear about ``\textit{how does the ego vehicle interact with surrounding vehicles over time in a specific space domain?}'' and ``\textit{how many interaction patterns exist?}.''

{\begin{figure}[t]
    \centering
    \includegraphics[width=1\linewidth]{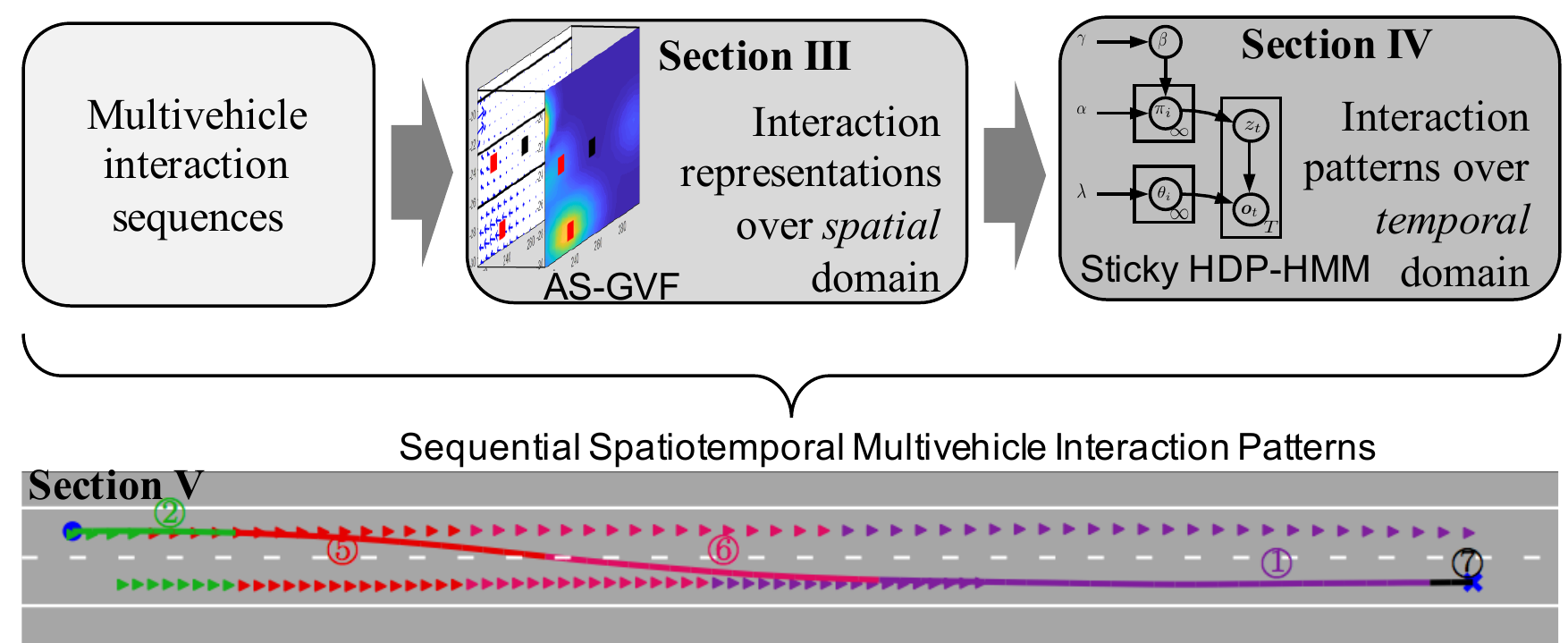}
    \caption{The framework of learning spatiotemporal interaction patterns during lane-change scenarios.}
    \label{fig_framework}
\end{figure}}

Motivated by the above two questions, this paper aims to develop an unsupervised learning framework (see Fig.\ref{fig_framework}) to figure out the underlying lane-change interaction patterns for potential decision-making applications. To this end, we will overcome two key challenges in
\begin{itemize}
\item Learning representation. The representation should capture the interactions of the ego vehicle with the nearby traffic agents while being insensitive to the number of agents in the environment.
\item Extracting interpretable interaction patterns. The number of patterns should increase adaptively when new scenarios are incrementally available.
\end{itemize}

Our main contributions are threefold:
\begin{enumerate}
    \item Proposing a general framework that leverages continuous and discrete stochastic processes to learn and recognize the lane-change interactions of the ego vehicle with its surrounding traffics on highways.
    \item Developing an acceleration-sensitive Gaussian velocity field to capture the interactions of the ego vehicle with its surrounding traffics in the \textit{spatial} space while reflecting the human drivers' intent. The dimensions of this field are invariant to the number of vehicles in the environment.
    \item Introducing a nonparametric approach to learn interaction patterns in the \textit{temporal} space automatically without any prior knowledge of the number of underlying patterns. 
\end{enumerate}


The remainder of this paper is organized as follows. Section \ref{sec2} reviews the related state-of-the-art. Section \ref{sec3} introduces the representation learning of multi-vehicle interactions. Section \ref{sec4} describes the methodolgy (i.e., Bayesian nonparametric learning) to extract interaction patterns, and Section \ref{sec5} explicitly analyzes the experimental results. Finally, conclusions are presented in Section \ref{sec6}.

\section{Related Works}\label{sec2}
\subsection{Lane-Change Behavior Analysis}
For driving behavior analysis and classification, there is no unified framework and theory to achieve them. The required methods and features rely on research motivations, such as controller design and decision-making. For instance, Woo {\it et al}.\cite{woo2017lane} defined four intentions (i.e., keeping, changing, arrival, and adjustment) to represent the procedure of changing lanes based on the distance from the centerline, the lateral velocity, and the potential feature. Yang {\it et al}.\cite{yang2018time} grouped the decision-making data into three driving styles - moderate, vague, and aggressive -- using the relative information between the ego vehicle and the target vehicle. Liu {\it et al}.\cite{liu2015classification} classified the driving maneuvers into moving forward, turning left, and changing lane for abnormal behavior detection based on the target vehicles' basic dynamic and kinematic states. The above behavioral categories were defined subjectively, which requires prior knowledge of different kinds of lane-change behaviors. However, prior knowledge is diverse over data analysts. Deo {\it et al}.\cite{deo2018would} empirically defined ten maneuver classes for freeway traffic by classifying the relative trajectories, including four lane-pass maneuvers, two overtake maneuvers, two cut-in maneuvers and two drifting-into-ego-lane maneuvers. However, Claussmann {\it et al}.\cite{claussmann2019review} distinguished them into eight common driving situations. Do {\it et al}.\cite{do2017human} segmented lane-change maneuvers of the ego vehicle into two stages over temporal space, and then listed $32$ occupancy grid states of right lane-change over spatial space, and finally grouped them into four categories. However, they assumed that the traffic agents on its left (right) neighboring lane would not impact the right (left) lane-change behavior. With the manually-predefined prior knowledge, Li {\it et al}.\cite{guangyu2019dbus} categorized lane change into nine driving actions based on GPS/IMU data and eight driving intentions based on front-view videos. 

The above discussion concludes that manually setting a reasonable number of clusters of behavioral interactions from real-world traffic data and then tagging these clusters is a tricky problem. Even if the number of clusters is already subjectively determined according to some specific semantic settings, the unsupervised clustering results may not always match the prior setting's semantic information \cite{habibi2019incremental}.  Our developed model aims at adaptively learning the lane-change interaction patterns of the ego vehicle with surrounding vehicles, but it does not need prior knowledge of the number of patterns.

\subsection{Representation Learning for Multi-Vehicle Interactions}

Representation construction for multi-vehicle interactions at each moment is critical to task performance but challenging due to the behavioral uncertainty of surrounding human-related agents. Human drivers can quickly and efficiently make decisions in complex, multidimensional environments because of the rugged ability to abstract representations of the environment \cite{niv2019learning}. In order to retrieve useful information from environments and make proper decisions for changing lanes while interacting with other vehicles efficiently, a variety of representatives were selected in state-of-the-art such as relative distance/speed/acceleration between the ego vehicle and the target vehicle \cite{yang2018time}. Besides, the deviations from the lane center and lateral and longitudinal velocity were also adopted in \cite{lienke2019predictive}. Moreover, the binary-state occupancy cells attached to the ego vehicle were used to represent surrounding scenarios \cite{kasper2012object, do2017human}. Some works were trying to figure out the influence of feature selection on lane-change behavior recognition and decision-making, thus offering  clues of selecting and constructing efficient features\cite{li2018effects, leonhardt2017feature}. The features mentioned above and their combinations can be directly measured using sensors; however, the dimension of all selected features is sensitive to the number of traffic agents in the environment.  

To make problem formulation tractable, researchers would classify the environment into several distinct groups according to the number of involved vehicles and select associated features subjectively. Lienke {\it et al}. \cite{lienke2019predictive} and Wang {\it et al}.\cite{wang2018extracting} limited the number of surrounding vehicles to six and five, respectively, which fails to handle the situations where the number is time-variant. Hu {\it et al}. \cite{hu2018probabilistic} used the features of the closest three vehicles around the ego vehicle to predict vehicle motion and intention, but the closest vehicles may be different at other timestamps, thus causing discontinuity of some features such as the relative positions. 

 One solution to handle an uncertain number of traffic agents is to model the pairwise interactions between the ego vehicle and each surrounding object independently. For instance, Xu {\it et al}.\cite{xu2017ego} selected frequent pairwise sequences as the data source and segmented the ROI into regional clusters, then modeled the spatial transition between regional clusters. Deo {\it et al}.\cite{deo2018would} treated the surrounding vehicles individually and established a probabilistic trajectory prediction model. However, considering the pairwise interactions may work well when with only several surrounding objects, but may have undesirable results when it comes to crowded and complex situations. Besides, pairwise trajectories can be similar for different traffic scenarios. Furthermore, the ego vehicle's decision-making is jointly affected by multiple surrounding objects other than pairwise interactions. Thus, these limitations require us to construct general features to represent the interactions among multiple vehicles jointly.

As widely selected features, potential fields have attracted many research interests in modeling multi-agent traffic scenarios due to its powerful capability to embrace different traffic factors friendly such as the types of road users, obstacles, and traffic regulations. Several kinds of potential fields have been developed for specific applications. For example, researchers in \cite{wolf2008artificial,wang2019crash} developed an artificial potential field considering different factors such as lane marker, road condition, and vehicle states. In \cite{rasekhipour2016potential}, a potential field integrated with vehicle dynamics was proposed to build a safe, robust path-planning controller for autonomous vehicles. Woo {\it et al}. developed \cite{woo2016dynamic} a dynamic potential field that is adaptable to the number of adjacent vehicles to predict lane change behavior. The driving safety field of describing driver-vehicle-road interactions was also proposed for collision warning \cite{wang2016drivingssafety}. More historical developments of field-relevant features refer to the review work \cite{claussmann2019review}. In this paper, we will develop a novel potential field based on Gaussian processes, which can deal a varying number of surrounding vehicles while capturing the dynamic interactions among multiple vehicles and encoding each surrounding vehicle's motion trends. More details will be provided in Section III. 

\subsection{Bayesian Nonparametrics for Driving Behavior}
Chopping complex sequential behavior into small segments and clustering them into groups helps us gain an insight into what happens inside it. Setting a reasonable number of clusters is always tricky in conventional clustering algorithms such as $k$-means and Gaussian mixture models (GMM)\cite{wang2019learning}. It requires trials and errors to get satisfied \cite{pelleg2000x}.  Nonparametric models, especially Bayesian nonparametric models, can outperform the traditional methods on these tasks\cite{birrell2014analysis}. Instead of treating the number of components and clusters as a constant, Bayesian nonparametric models treat it as a random variable by adding a functional distribution layer, for example, by adding a Dirichlet process to GMM\cite{rasmussen2000infinite,gorur2010dirichlet}. Bayesian nonparametric has demonstrated its powerful ability to model and predict dynamic processes with the number of patterns a  priori unknown\cite{fox2007sticky,raman2016activity,jochmann2015modeling}.  

In driving behavior analysis and prediction, the Bayesian nonparametric models have also been implemented. Researchers introduced the Dirichlet process (DP) as a prior over the number of driving behavior patterns formulated by a hidden Markov model (HMM). Hamada {\it et al}.\cite{hamada2016modeling} utilized beta process autoregressive HMM to segment driving behaviors into different states. Wang {\it et al}.\cite{wang2018extracting, wang2018driving} adopted a hierarchical DP-HMM to analyze the driving styles using multi-dimensional time-series driving data. Mahjoub {\it et al}.\cite{mahjoub2018stochastic} established a framework based on hierarchical DP to jointly model the driving behavior through forecasting the vehicle dynamical time-series. The theoretical basis of discrete Bayesian nonparametrics will be introduced in Section IV. This paper combines the continuous and discrete Bayesian nonparametrics to learn interaction patterns during lane change from high-dimensional time series.

\section{Spatial Representation Learning of Multi-Vehicle Interactions}\label{sec3}
Efficient representations of sequential interactions are critical to the performance of tasks \cite{niv2019learning} but challenging in the real-world as many time-variant factors exist in dynamic scenarios such as the number of vehicles. Human brains can interact with their surroundings by actively forming a field of specific physiological/perceptual measures that reflect the behavioral relevance of a stimulus to actions\cite{bufacchi2018action}. 
Inspired by the above, the relevance between the driving environments and the driver's actions can reveal the human driver's interaction with other agents in the perception area. In order to construct this kind of representation, our previous work\cite{zhang2019general} introduced a Gaussian process regression model to estimate the influences of each vehicle's velocity at any location in the environment, termed as velocity fields. 
In \cite{zhang2019general}, the authors assume that each vehicle has a symmetric impact on its nearby region, ignoring the surrounding vehicles' intent, such as acceleration and deceleration. However, the brake action (deceleration) of a car would make a stronger impact on its behind area than its front area, and the left-turning maneuver would have a broader impact range on the left than on the right. In this section, we will first briefly revisit the model in our previous work\cite{zhang2019general} and then extend it to an acceleration-sensitive Gaussian velocity field by accounting for driver intents' impacts.

\begin{figure}[t]
\centering
\includegraphics[width = 0.95\linewidth]{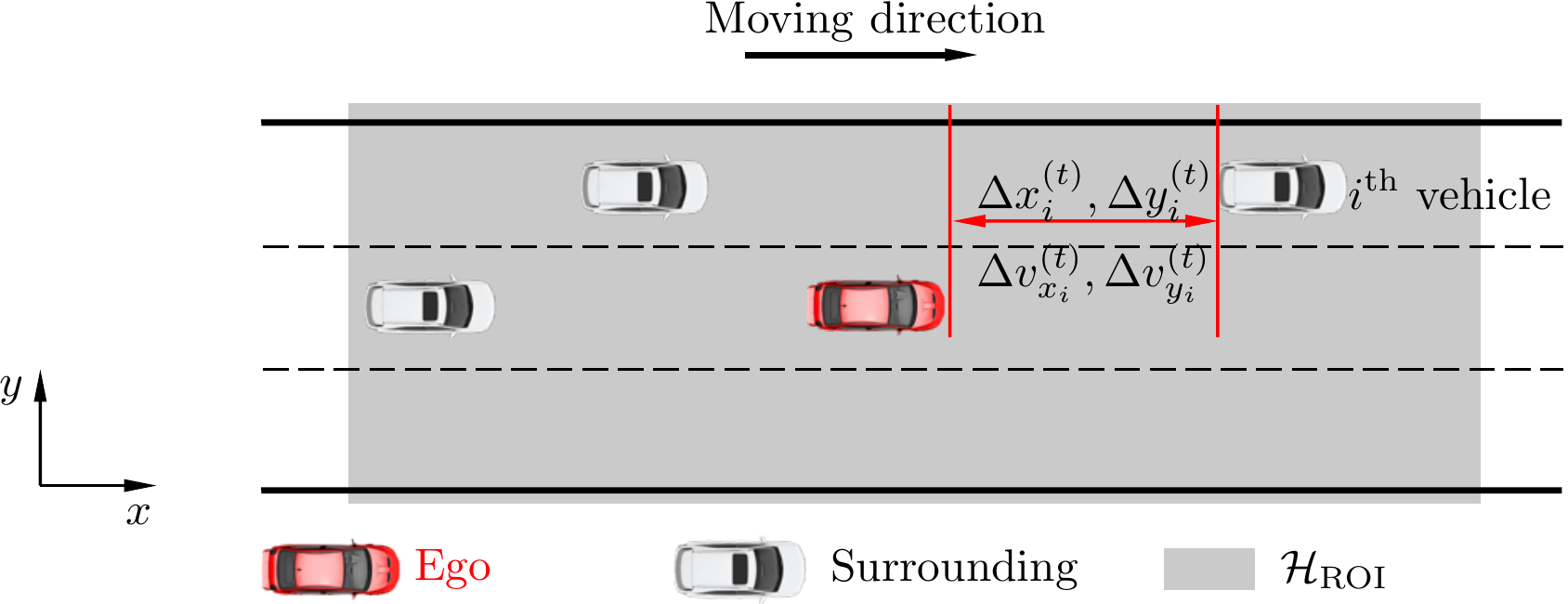}
\caption{Illustration of lane-change scenarios.}
\label{fig:lanechangescenario}
\end{figure}

\subsection{Gaussian Process and Velocity Fields}
The surrounding vehicles in the predefined ROI, $\mathcal{H}_{\mathrm{ROI}}$, would impact lane-change decisions and maneuvers of the ego vehicle. Unlike existing works that aim to develop controllers and predict lane change behaviors\cite{rasekhipour2016potential,woo2016dynamic,wang2016drivingssafety,zhu2019probabilistic,qin2019recurrent}, our focus is on capturing the interactions of the ego vehicle with other vehicles in dynamic environments. The ego vehicle will assess the interactions using their relative velocity against positions, rather than remain stationary to surrounding obstacles during the entire lane-change procedure.

Given $N_{t}$ surrounding vehicles at time $t$ ($N_{t}$ is time-variant) in the environment $\mathcal{H}_{\mathrm{ROI}}$ (see Fig. \ref{fig:lanechangescenario}), the set of their location and relative velocity to the ego vehicle is denoted by $\{(x_{n}^{(t)}, y_{n}^{(t)})\}_{n=1}^{N_{t}}$ and $\{(\Delta v_{x_n}^{(t)}, \Delta v_{y_n}^{(t)})\}_{n=1}^{N_{t}}$, respectively. Our goal is to estimate the distribution of the relative velocity over any location in $\mathcal{H}_{\mathrm{ROI}}$. To simplify the model, we assume that the distributions in the directions of $x$ and $y$ are independent. The estimation in either direction can be specified by Gaussian process regression (GPR). Thus, combining the distributions at two directions forms a Gaussian velocity field (GVF) over $\mathcal{H}_{\mathrm{ROI}}$. More backgrounds refer to Chapter II in \cite{rasmussen2003gaussian}. To help readers catch up on the basic concept of the GVF, we first revisit the GPR for prediction with noise-free observations. 

\subsubsection{GPR}

\begin{figure}[t]
\centering
\includegraphics[width = \linewidth]{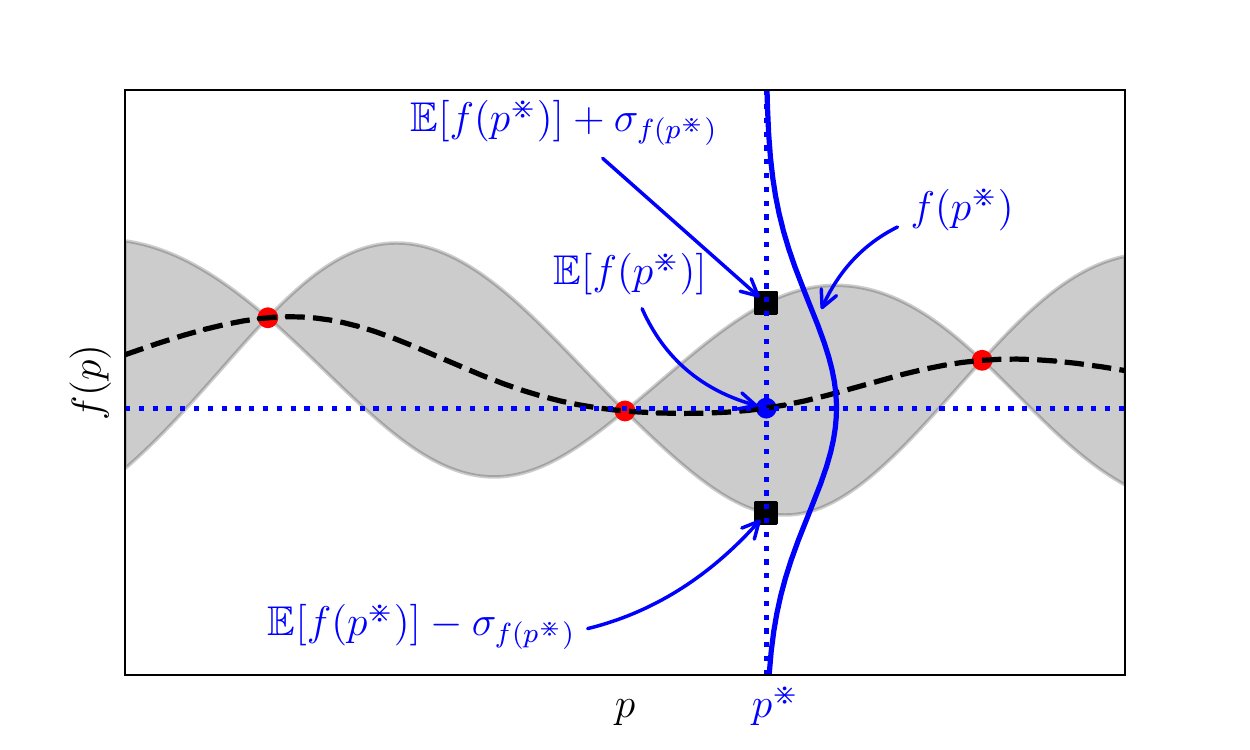}
\caption{  Illustration of GPR in one dimension. Red dots are the observations $\{p, f(p)\}$. $p$ and $f(p)$ are the measure space of inputs and outputs, respectively. Black dash line represent the expectation of $f(p)$, i.e., $\mathbb{E}[f(p)]$, for all $p$. The gray-shaded area represent the region $\mathbb{E}[f(p)] \pm \sigma_{f(p)}$.}
\label{fig:gpr}
\end{figure}

As the above discussion, the velocity distributions at each location of $\mathcal{H}_{\mathrm{ROI}}$ in the $x$ and $y$ directions are estimated independently. Hence, each direction's GPR model is with two-dimensional inputs (i.e., location ($x,y$)) and one-dimensional output (i.e., $\Delta v_{x}$ or $\Delta v_{y}$). For simplification, we take a Gaussian process with one-dimensional  input for illustration (see Fig. \ref{fig:gpr}), which is defined as a probability distribution over function $f(p)\in \mathbb{R}$ (Note that $f(p)$ is a random variable for a specific $p$) and $p$ is the input. According to the definition of Gaussian process, any finite number of collections of random variables $f(\cdot)$ still have a joint Gaussian distributions; for instance, any finite functional collections over $p$ and $p^{\prime}$ have a Gaussian distribution
\begin{equation}
f(p)\sim \mathcal{N}(\mu (p),k(p,p^{\prime}))
\end{equation}
with the mean function $\mu(p)$ and the covariance function $k(p,p^{\prime})$\footnote[1]{The covariance function $k(p,p^{\prime})$ can be treated as the simplified form of $\mathrm{cov}(f(p),f(p^{\prime}))$ -- the covariance of functional random variables $f(p)$ and $f(p^{\prime})$.}, computed by
\begin{subequations}
\begin{align}
    \mu (p)&=\mathbb{E}[f(p)] \\
    k(p,p^{\prime})&=\mathbb{E}[(f(p)-\mu (p))(f(p^{\prime})-\mu (p^{\prime}))] \label{kernel_onedim}
\end{align}
\end{subequations}
Note that when $p = p^{\prime}$, $k(p, p^{\prime}) = \sigma_{f(p)}$ is the standard variance of the functional variable $f(p)$, as shown in Fig. \ref{fig:gpr}. 

Assuming that we have the observation set $\{p_{n},f(p_{n})\}_{n=1}^{N}$, our aim is to estimate the distribution of $\{f(p^{\ast}_{m})\}$ over any finite collections of possible inputs $\{p^{\ast}_{m}\}_{m=1}^{M}$. For convenience, we denote $\mathbf{p} = [p_{1}, \dots, p_{N}]$, $\mathbf{f} = [f(p_{1}), \dots, f(p_{N})]$, $\mathbf{p}^{\ast} = [p_{1}^{\ast}, \dots,  p_{M}^{\ast}]$, and $\mathbf{f}^{\ast} = [f(p_{1}^{\ast}), \dots, f(p_{M}^{\ast})]$. According to the properties of Gaussian process, both $\mathbf{f}$ and $\mathbf{f}^{\ast}$ are the instantiation of the finite collections of random variables and have Gaussian distributions. The prior knowledge about the mean of $\mathbf{f}$ is usually unavailable; therefore, we set the prior distributions of $\mathbf{f}$ and $\mathbf{f}^{\ast}$ with zero-mean value as
\begin{subequations}
\begin{align}
    \mathbf{f}&\sim \mathcal{N}(\boldsymbol{0},\mathbf{K}(\mathbf{p},\mathbf{p}))\\
    \mathbf{f^{\ast}}&\sim \mathcal{N}(\boldsymbol{0},\mathbf{K}(\mathbf{p}^{\ast},\mathbf{p}^{\ast}))
\end{align}
\end{subequations}
where $\mathbf{K}$ is the covariance matrix with elements $k_{ij}=k(p_i,p_j)$. Given the noise-free training data $\{p_i,f(p_i)\}$, the joint distribution of the training outputs $\mathbf{f}$ and the test outputs $\mathbf{f}^{\ast}$ still has a zero-mean Gaussian distribution with
\begin{equation}\label{eq:gaussiandistribution}
\begin{bmatrix}
\mathbf{f} \\ \mathbf{f}^{\ast}
\end{bmatrix}
\sim
\mathcal{N}\left(\mathbf{0},
\begin{bmatrix}
\mathbf{K}(\mathbf{p},\mathbf{p}), & \mathbf{K}(\mathbf{p},\mathbf{p}^{\ast}) \\ \mathbf{K}(\mathbf{p}^{\ast},\mathbf{p}), & \mathbf{K}(\mathbf{p}^{\ast},\mathbf{p}^{\ast})
\end{bmatrix}
\right)
\end{equation}
with $\mathbf{K}(\mathbf{p},\mathbf{p})\in\mathbb{R}^{N\times N}$, $\mathbf{K}(\mathbf{p},\mathbf{p}^{\ast})\in \mathbb{R}^{N\times M}$, $\mathbf{K}(\mathbf{p}^{\ast},\mathbf{p})\in \mathbb{R}^{M\times N}$, and $\mathbf{K}(\mathbf{p}^{\ast},\mathbf{p}^{\ast})\in \mathbb{R}^{M\times M}$.

We aim to draw samples from the posterior distribution that contains the training data information, rather than draw random functions directly from the prior. According to the multivariate conditional Gaussian distribution with (\ref{eq:gaussiandistribution}), the distribution of test outputs conditioning on the training outputs are given by
\begin{equation}
\begin{split}
    \mathbf{f}^{\ast}|\mathbf{p}^{\ast},\mathbf{p},\mathbf{f} \sim \mathcal{N}(\boldsymbol{\mu}_{\mathbf{f}^{\ast}}, \boldsymbol{\Sigma}_{\mathbf{f}^{\ast}})
\end{split}
\end{equation}
with mean $\boldsymbol{\mu}_{\mathbf{f}^{\ast}}$ and covariance $\boldsymbol{\Sigma}_{\mathbf{f}^{\ast}}$ as
\begin{subequations}
\begin{align}
    \boldsymbol{\mu}_{\mathbf{f}^{\ast}}&=\mathbf{K}(\mathbf{p}^{\ast},\mathbf{p})\mathbf{K}(\mathbf{p},\mathbf{p})^{-1}\mathbf{f} \label{eq:mean}\\
    \boldsymbol{\Sigma}_{\mathbf{f}^{\ast}}&=\mathbf{K}(\mathbf{p}^{\ast},\mathbf{p}^{\ast})-\mathbf{K}(\mathbf{p}^{\ast},\mathbf{p})\mathbf{K}(\mathbf{p},\mathbf{p})^{-1}\mathbf{K}(\mathbf{p},\mathbf{p}^{\ast})
\end{align}
\end{subequations}
 Interested readers who are not familiar with the conditional Gaussian distribution and Gaussian process can refer to Chapter 2.3.1 and Chapter 6.4 in \cite{bishop2006pattern} for more details of deriving these equations. Therefore, given any input $p$, the distribution of functional random variable $f(p)$ of any other possible input $p^{\ast}$ over the input space can be computed directly.

\subsubsection{Gaussian Velocity Fields (GVF)}
To describe the probability of multi-vehicle interactions at any location in a predefined ROI, we introduce the GPR to compute their estimates. The relative velocities specify the interaction at any location against their relative positions. For computational efficiency, the estimated relative velocity in the $x$ and $y$ directions at a single location is independent of each other. In what follows, we compute the interaction in the $x$-direction, which is also applicable for estimation in the $y$-direction. 

Given the training data $\{(\Delta v_x(p_i),\Delta v_y(p_i))\}_{i=1}^{N_{t}}$ at associated locations $\mathbf{p} = \{p_{i}\}= \{(x_i,y_i)\}_{i=1}^{N_{t}}\in \mathcal{H}_{\mathrm{ROI}}$ at time $t$, all the functional random variables over $\mathcal{H}_{\mathrm{ROI}}$ are Gaussian  processes. Thus, we can compute the distribution of $\Delta v_{x}$ at a finite number of collections of any possible locations $\mathbf{p}^{\ast}$ by
\begin{equation}
\Delta v_{x}(\mathbf{p}^{\ast}) \sim \mathcal{N}(\boldsymbol{\mu}_{x} (\mathbf{p}^{\ast}), \boldsymbol{\Sigma}_{x}(\mathbf{p}^{\ast})), \ \forall \mathbf{p}^{\ast} \in \mathcal{H}_{\mathrm{ROI}}
\end{equation}
where $\boldsymbol{\mu}_{x} (\mathbf{p}^{\ast})$ is the mean and $\mathbf{\Sigma}_{x}(\mathbf{p}^{\ast})$ is the covariance. Unlike (\ref{kernel_onedim}) in one dimension, each element of $\mathbf{\Sigma}_{x}(\mathbf{p}^{\ast})$ is specified by a multivariate kernel $k(p_{i}, p_{j})$ with $p_{i}, p_{j} \in \mathbf{p}^{\ast}$, and $k$ is defined as the standard squared exponential function
\begin{equation}\label{eq:kernel}
k(p_i, p_j) = A \exp\left({-\frac{(x_i-x_j)^2}{2\sigma ^2_x} - \frac{(y_i-y_j)^2}{2\sigma ^2_y}}\right)
\end{equation}
where $A$ is an amplitude constant, $\sigma_x$ and $\sigma_y$ are length-scale constant controlling how the correlations are decay with respect to distance. A higher value of $\sigma_{x}$ (or $\sigma_{y}$) indicates slow decay, and thus farther points will have near-zero covariance or correlations. The standard squared exponential kernel is selected, such that the nearby relative velocity samples are more correlated than the one farther away in the field of $\mathcal{H}_{\mathrm{ROI}}$.

Therefore, according to (\ref{eq:mean}), given the sets of relative velocity $\Delta \mathbf{V}_{x} = \{\Delta v_{x}(p_{i})\}_{i=1}^{N_{t}}$ at locations $\mathbf{p}$ of all surrounding vehicles as the training data, the posterior distribution of the predicted relative velocities $\Delta v_x(\mathbf{p}^{\ast})$ at any location $\mathbf{p}^{\ast}$ in the ROI can be efficiently computed via
\begin{equation}\label{eq2}
\boldsymbol{\mu}_{x}(\mathbf{p}^{\ast}) = \mathbf{K}(\mathbf{p}^{\ast}, \mathbf{p}) \mathbf{K}(\mathbf{p}, \mathbf{p})^{-1} \Delta \mathbf{V}_{x}, \ \forall \mathbf{p}^{\ast} \in \mathcal{H}_{ROI}
\end{equation}
Similarly, we can obtain the distribution of relative velocity at a finite collection of any location $\mathbf{p}^{\ast}$ in the $y$-direction, denoted as $\boldsymbol{\mu}_{y}(\mathbf{p}^{\ast})$.

After computing the expectation of the relative velocity in the $x$ and $y$ directions independently, we can estimate the value and direction of the relative velocity at any location in the ROI.

\subsection{Acceleration-Sensitive GVF (AS-GVF)}
The above section describes multi-vehicle interactions based on the GVF specified by the ego vehicle's relative velocity to other agents in the surrounding area. However, equation (\ref{eq:kernel}) assumes that each vehicle has a symmetrical impact on their surroundings, which leads to the deviation from the real traffic situation as humans are also active to stimulus (e.g., acceleration) other than proximity \cite{bufacchi2018action}. For example, a nearby moving vehicle would draw a non-symmetric stimulus on its surrounding space for the human drivers due to maneuvers such as deceleration and acceleration \cite{moridpour2015impact}. To make the distance-based GVF discussed above capable of capturing the vehicle motion inertia, we revise the symmetric and isotropic kernel of (\ref{eq:kernel}) by integrating the acceleration of surrounding vehicles.%

The modified velocity field is expected to reflect  each surrounding vehicle's driving intent and instant stimulus in $\mathcal{H}_{\mathrm{ROI}}$. To this end, 
we reconstructed (\ref{eq2}) by integrating the acceleration into $\mathbf{K}$ through a skewed matrix $\mathbf{K}^{\prime}$, thus obtaining
\begin{equation}\label{eq4}
\begin{split}
\boldsymbol{\mu}^{\prime}_{x}(\mathbf{p}^{\ast}) 
 & = \underbrace{\mathbf{K}^{\prime}(\mathbf{p}^{\ast},\mathbf{p},\mathbf{a_x},\mathbf{a_y})\circ\mathbf{K}(\mathbf{p}^{\ast}, \mathbf{p})}_{\mathrm{Hadamard \ product}} \mathbf{K}(\mathbf{p}, \mathbf{p})^{-1} \Delta \mathbf{V}_{x}
\end{split}
\end{equation}
where ($\circ$) represents the Hadamard product, $\mathbf{a_x}$ and $\mathbf{a_y}$ are the set of accelerations in the $x$ and $y$ directions, respectively. The skewed matrix is
\begin{equation*}
\mathbf{K}^{\prime} = 
\begin{bmatrix}
k^{\prime}_{1,1} & \cdots & k^{\prime}_{1,N_{t}}\\
\vdots & \ddots & \vdots \\
k^{\prime}_{M,1} & \cdots & k^{\prime}_{M,N_{t}}
\end{bmatrix}_{M\times N_{t}}
\end{equation*}
with $k^{\prime}_{ij}= k^{\prime}(p_{i}, p_{j}, \mathbf{a}_{j})$, $\mathbf{a}_{j} = [a_{j,x}, a_{j,y}]$, $j\in \{1, \dots, N_{t}\}$, and $i\in \{1, \dots, M\}$, where $N_{t}$ is the number of vehicles in the environment at time $t$, and $M$ is the number of testing locations in the $x$ (or $y$) direction over $\mathcal{H}_{\mathrm{ROI}}$. Therefore, the skew matrix is in $\mathbb{R}^{M\times N_{t}}$. In order to make $k^{\prime}_{ij}$ practically feasible, two constraints of acceleration should be satisfied:

\begin{itemize}
\item {A vehicle heading straight with small acceleration would have an almost symmetric influence over the environment. As the acceleration approaches to zero, the value of AS-GVF would converge to GVF.}

\item An aggressive acceleration (deceleration) would lead to a high (low) field tensity in the (opposite) direction of the acceleration.
\end{itemize}
Therefore, $k^{\prime}_{ij}$ should be a function of the relative position and acceleration and provide a high value in the accelerating direction. Here, we design $k^{\prime}$ by multiplying two logistic functions in the direction of $a_x$ and $a_y$ as
\begin{align}
    k^{\prime}(p_i,p_j,\mathbf{a}_{j})=\prod_{\ell=x, y}\frac{\xi_\ell}{1+e^{-\lambda_\ell a_{j,\ell}(\ell_i-\ell_j)}}
\end{align}
where $\xi_\ell$ is the normalization factor and $\lambda_\ell$ is the skewing factor.

\subsection{Dimension Reduction of AS-GVF}

The developed AS-GVF captures the multi-vehicle interactions using the estimated relative velocity at each location in the environment, formed as a tensor. However, a high-dimensional tensor makes it computationally infeasible to implement Bayesian inference with sampling algorithms. Therefore, reducing the high-dimensional tensor into a low dimension is a prerequisite for efficient recognition of interaction patterns. 

\begin{figure*}[t]
    \centering
    \includegraphics[width=\textwidth]{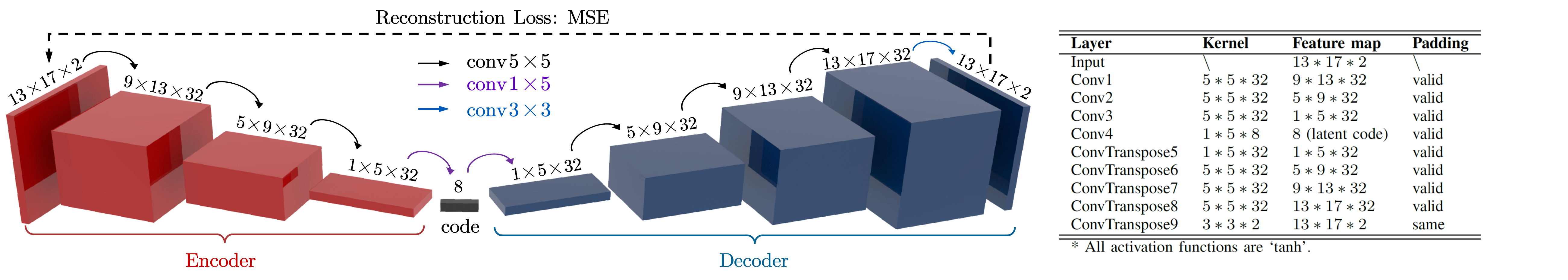}
    \caption{Architecture of the convolutional autoencoder and its settings.}
    \label{fig_CAE}
\end{figure*}

This paper implements an unsupervised technique -- convolutional autoencoder (CAE) to generate latent codes to represent AS-GVF at each time. A typical autoencoder is composed of an encoder $h=g_{1}(s)$ and a decoder $r=g_{2}(h)$. It reconstructs its inputs $s$ and sets the target values of output $\hat{s}=g_{2}(g_{1}(s))$ to be equal to $s$, where $g_{1}(\cdot )$ and $g_{2}(\cdot)$ are the activation functions\cite{ng2011sparse}. 
The tensor of AS-GVF characterizes the spatial interactions of vehicles in the environment; thus a convolutional operation over the velocity field can take the satisfying performance of feature extraction. In our case, the CAE is treated as a deep autoencoder with fully convolutional layers, as shown in Fig. \ref{fig_CAE}.
\section{Spatiotemporal Interaction Patterns Learning}\label{sec4}
The AS-GVF delivers the feature representatives of multi-vehicle interactions at each frame. Our goal is to cluster the sequential interactions into groups according to the similarity. However, it is infeasible to manually set the number of interaction patterns due to data increases. This section will describe a discrete Bayesian nonparametric approach to segment the behavioral sequences into fundamental pieces, called primitives \cite{zhang2019general,zhang2019learning} while clustering them into groups. We first revisit the basic concept of primitives and then introduce the related Bayesian nonparametric methods to learn them. 

\subsection{Primitives and Interaction Patterns}
Human skills consist of primitive information processing elements, and several of these primitive elements are necessary for even a single step in a task \cite{taatgen2013nature}. Besides, segmenting complex interactive behavioral sequences into small recognizable elements gives an insight into driving behavior recognition and prediction\cite{hamada2016modeling,taniguchi2015sequence}. Supported by these conclusions, the multi-vehicle sequential interactions during lane change can be segmented into a bunch of cascading fundamental blocks (i.e., primitives). More specifically, the primitives in this paper are referred to as the segments of the sequential interaction behavior during lane change over time. Therefore, each type of primitives represents a basic lane-change interaction pattern, and a complex lane-change scenario consists of several types of traffic primitives. For the sake of unification, we will refer to traffic primitives as interaction patterns in the following.

\subsection{Bayesian Nonparametric Models}
The number of interaction patterns during lane change, intuitively, should increase as more new scenarios being explored. This kind of functionality can be achieved with nonparametric models whose parameters will theoretically increase with more observed data. This paper implements a nonparametric model in a Bayesian framework with an infinite-dimensional parameter space. Besides, the sequential interactions has the Markov property during the lane-change procedure, which makes it reasonable to employ a hidden Markov model (HMM) to formulate the dynamic transition of these interaction patterns. Another expected capability of Bayesian nonparametric models is grouping these patterns while taking segmentation, which could be carried out by introducing a hierarchical Dirichlet process (HDP) over HMM, called HDP-HMM\cite{fox2011sticky}. In what follows, we will revisit the relevant theoretical preliminaries.

\subsubsection{Hidden Markov Model}

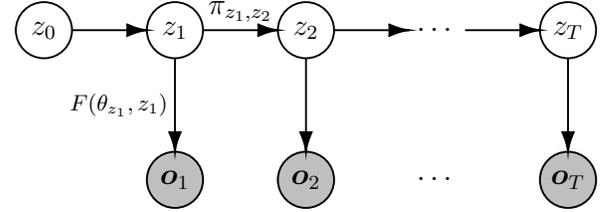
\begin{figure}[t]
\centering
\resizebox{0.9\linewidth}{!}{\tikzset{every picture/.style={line width=0.75pt}} 

\begin{tikzpicture}[x=0.75pt,y=0.75pt,yscale=-1,xscale=1]

\draw [black] (130,100) circle(15);
\draw (145, 100) -- (185, 100);
\filldraw [shift={(185,100)}, rotate = 180] [color={rgb, 255:red, 0; green, 0; blue, 0 }  ][line width=0.75]    (10.93,-3.29) .. controls (6.95,-1.4) and (3.31,-0.3) .. (0,0) .. controls (3.31,0.3) and (6.95,1.4) .. (10.93,3.29);

\draw [black] (200,100) circle(15);
\draw (215, 100) -- (255, 100);
\filldraw [shift={(255,100)}, rotate = 180] [color={rgb, 255:red, 0; green, 0; blue, 0 }  ][line width=0.75]    (10.93,-3.29) .. controls (6.95,-1.4) and (3.31,-0.3) .. (0,0) .. controls (3.31,0.3) and (6.95,1.4) .. (10.93,3.29);

\draw [black] (270,100) circle(15);
\draw (285, 100) -- (325, 100);
\filldraw [shift={(325,100)}, rotate = 180] [color={rgb, 255:red, 0; green, 0; blue, 0 }  ][line width=0.75]    (10.93,-3.29) .. controls (6.95,-1.4) and (3.31,-0.3) .. (0,0) .. controls (3.31,0.3) and (6.95,1.4) .. (10.93,3.29);

\draw [black] (410,100) circle(15);
\draw (355, 100) -- (395, 100);
\filldraw [shift={(395,100)}, rotate = 180] [color={rgb, 255:red, 0; green, 0; blue, 0 }  ][line width=0.75]    (10.93,-3.29) .. controls (6.95,-1.4) and (3.31,-0.3) .. (0,0) .. controls (3.31,0.3) and (6.95,1.4) .. (10.93,3.29);

\filldraw [gray!50] (200,180) circle(15);
\draw (200,180) circle(15);
\draw (200, 115) -- (200, 165);
\filldraw [shift={(200,165)}, rotate = -90] [color={rgb, 255:red, 0; green, 0; blue, 0 }  ][line width=0.75]    (10.93,-3.29) .. controls (6.95,-1.4) and (3.31,-0.3) .. (0,0) .. controls (3.31,0.3) and (6.95,1.4) .. (10.93,3.29);

\filldraw [gray!50] (270,180) circle(15);
\draw (270, 180) circle(15);
\draw (270, 115) -- (270, 165);
\filldraw [shift={(270,165)}, rotate = -90] [color={rgb, 255:red, 0; green, 0; blue, 0 }  ][line width=0.75]    (10.93,-3.29) .. controls (6.95,-1.4) and (3.31,-0.3) .. (0,0) .. controls (3.31,0.3) and (6.95,1.4) .. (10.93,3.29);

\filldraw [gray!50] (410,180) circle(15);
\draw (410, 180) circle(15);
\draw (410, 115) -- (410, 165);
\filldraw [shift={(410,165)}, rotate = -90] [color={rgb, 255:red, 0; green, 0; blue, 0 }  ][line width=0.75]    (10.93,-3.29) .. controls (6.95,-1.4) and (3.31,-0.3) .. (0,0) .. controls (3.31,0.3) and (6.95,1.4) .. (10.93,3.29);

\draw (340,100) node  [font=\large]  {$\cdots $};
\draw (340,180) node  [font=\large]  {$\cdots $};
\draw (130,100) node  [font=\large]  {$z_{0}$};
\draw (200,100) node [font=\large]   {$z_{1}$};
\draw (270,100) node [font=\large]   {$z_{2}$};
\draw (410,100) node  [font=\large]  {$z_{T}$};
\draw (200,180) node  [font=\large]  {$\boldsymbol{o}_{1}$};
\draw (270,180) node  [font=\large]  {$\boldsymbol{o}_{2}$};
\draw (410,180) node  [font=\large]  {$\boldsymbol{o}_{T}$};
\draw (235, 90) node [font=\large] {$\pi_{z_{1},z_{2}}$};
\draw (170, 140) node [font=\small] {$F(\theta_{z_{1}}, z_{1})$};

\end{tikzpicture}}
\caption{The graphical illustration of hidden Markov models.}
\label{fig6}
\end{figure}

The lane-change interaction behavior is treated as a certain cascade of sequential interaction patterns whose dynamic process can be modeled as a probabilistic inferential process. Specifically, we model the dynamics of interaction patterns over the time horizon based on the HMM structure consisting of two parts (see Fig. \ref{fig6}): discrete latent interaction patterns (denoted by non-shaded nodes $z_t$) and observed interaction representations (denoted by shaded nodes $\boldsymbol{o}_t$). Our learning task aims to infer the latent pattern from the observed interaction representations. In HMM, each observed interaction representation $\boldsymbol{o}_{t}$ at time $t$ would be assigned an interaction pattern $z_{t} \in \mathcal{Z}$, where $\mathcal{Z}$ is the set of all types of interaction patterns. For convenient formulation, we specify each type of pattern in $\mathcal{Z}$ using an integer number, i.e.,  $z_{t}$ is equal to any single element in $\{1,2,\dots,|\mathcal{Z}|\}$, where $|\mathcal{Z}|$ is the size of the set $\mathcal{Z}$. 
The probability that the pattern at time $t$ transits to the pattern at time $t+1$ is denoted by $\pi_{z_{t}, z_{t+1}} $, which is shortened as $ \pi_{i,j} $ when $z_{t} = i$ and $z_{t+1} = j$ with $i,j\in \mathcal{Z}$. Thus,  $\pi = \{\pi_{i, j}\}_{i, j=1}^{|\mathcal{Z}|}$ is the probabilistic transition matrix, and $\pi_{i}$ represents the probability mass function (PMF) of interaction patterns conditioned pattern $i$. According to the above definition, given the current latent pattern $z_{t}$ and the emission parameter $\theta_{z_{t}}$, the observed representation $\boldsymbol{o}_t$ is drawn from a function $F(\boldsymbol{o}_t|\theta_{z_{t}},z_t)$ parameterized by $\theta_{z_{t}}$. Thus, the general form of HMM can be written as
\begin{subequations}
\begin{align}
z_t|z_{t-1} & \sim \pi_{z_{t-1}} \label{eq:latentpattern}\\
\boldsymbol{o}_t|z_t & \sim F(\theta_{z_t}, z_{t}) \label{eq:emission}
\end{align}
\end{subequations}
where $\pi_{z_{t-1}}$ represents the PMF of $z_{t}$ conditional on the one-step-back latent pattern $z_{t-1}$.
The classical treatments of HMM require specifying the size of $\mathcal{Z}$; however, $|\mathcal{Z}|$ is unknown a priori in our case, that is, we do not precisely know how many interaction patterns exist in a bunch of sequential lane-change behaviors. We introduce a discrete prior probability distribution in infinite-dimensional space based on Dirichlet Processes (DP) to solve this problem.

\subsubsection{DP and Hierarchical DP}

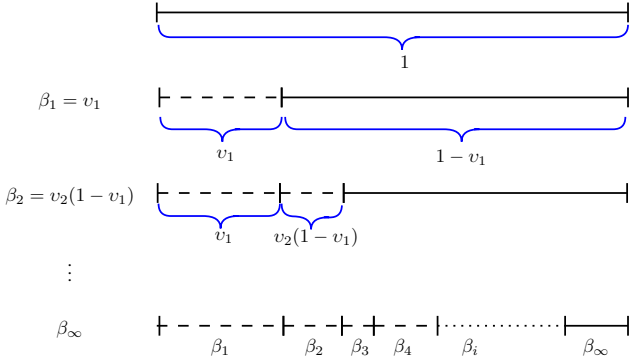
\begin{figure}[t]
\centering
\resizebox{0.97\linewidth}{!}{\tikzset{every picture/.style={line width=0.75pt}} 

\begin{tikzpicture}[x=0.75pt,y=0.75pt,yscale=-1,xscale=1]

\draw    (139,36) -- (405,36) ;
\draw [shift={(405,36)}, rotate = 180] [color={rgb, 255:red, 0; green, 0; blue, 0 }  ][line width=0.75]    (0,5.59) -- (0,-5.59)   ;
\draw [shift={(139,36)}, rotate = 180] [color={rgb, 255:red, 0; green, 0; blue, 0 }  ][line width=0.75]    (0,5.59) -- (0,-5.59)   ;
\draw  [color={rgb, 255:red, 0; green, 0; blue, 255 }  ,draw opacity=1 ] (140,43) .. controls (140,47.67) and (142.33,50) .. (147,50) -- (267.49,50) .. controls (274.16,50) and (277.49,52.33) .. (277.49,57) .. controls (277.49,52.33) and (280.82,50) .. (287.49,50)(284.49,50) -- (397.5,50) .. controls (402.17,50) and (405,47.67) .. (405,43) ;
\draw    (209.5,84) -- (405,84) ;
\draw [shift={(405,84)}, rotate = 180] [color={rgb, 255:red, 0; green, 0; blue, 0 }  ][line width=0.75]    (0,5.59) -- (0,-5.59)   ;
\draw [shift={(209.5,84)}, rotate = 180] [color={rgb, 255:red, 0; green, 0; blue, 0 }  ][line width=0.75]    (0,5.59) -- (0,-5.59)   ;
\draw  [color={rgb, 255:red, 0; green, 0; blue, 255 }  ,draw opacity=1 ] (141,95) .. controls (141,99.67) and (143.33,102) .. (148,102) -- (166.03,102) .. controls (172.7,102) and (176.03,104.33) .. (176.03,109) .. controls (176.03,104.33) and (179.36,102) .. (186.03,102)(183.03,102) -- (202.5,102) .. controls (207.17,102) and (209.5,99.67) .. (209.5,95) ;
\draw    (244.5,138) -- (405,138) ;
\draw [shift={(404.5,138)}, rotate = 180] [color={rgb, 255:red, 0; green, 0; blue, 0 }  ][line width=0.75]    (0,5.59) -- (0,-5.59)   ;
\draw [shift={(244.5,138)}, rotate = 180] [color={rgb, 255:red, 0; green, 0; blue, 0 }  ][line width=0.75]    (0,5.59) -- (0,-5.59)   ;
\draw  [dash pattern={on 4.5pt off 4.5pt}]  (140.5,84) -- (209.5,84) ;
\draw [shift={(209.5,84)}, rotate = 180] [color={rgb, 255:red, 0; green, 0; blue, 0 }  ][line width=0.75]    (0,5.59) -- (0,-5.59)   ;
\draw [shift={(140.5,84)}, rotate = 180] [color={rgb, 255:red, 0; green, 0; blue, 0 }  ][line width=0.75]    (0,5.59) -- (0,-5.59)   ;
\draw  [dash pattern={on 4.5pt off 4.5pt}]  (139.5,138) -- (208.5,138) ;
\draw [shift={(208.5,138)}, rotate = 180] [color={rgb, 255:red, 0; green, 0; blue, 0 }  ][line width=0.75]    (0,5.59) -- (0,-5.59)   ;
\draw [shift={(139.5,138)}, rotate = 180] [color={rgb, 255:red, 0; green, 0; blue, 0 }  ][line width=0.75]    (0,5.59) -- (0,-5.59)   ;
\draw  [dash pattern={on 4.5pt off 4.5pt}]  (208.5,138) -- (244.5,138) ;
\draw [shift={(244.5,138)}, rotate = 180] [color={rgb, 255:red, 0; green, 0; blue, 0 }  ][line width=0.75]    (0,5.59) -- (0,-5.59)   ;
\draw [shift={(208.5,138)}, rotate = 180] [color={rgb, 255:red, 0; green, 0; blue, 0 }  ][line width=0.75]    (0,5.59) -- (0,-5.59)   ;
\draw  [color={rgb, 255:red, 0; green, 0; blue, 255 }  ,draw opacity=1 ] (211.5,95) .. controls (211.5,99.67) and (213.83,102) .. (218.5,102) -- (301.09,102) .. controls (307.76,102) and (311.09,104.33) .. (311.09,109) .. controls (311.09,104.33) and (314.42,102) .. (321.09,102)(318.09,102) -- (396.5,102) .. controls (401.17,102) and (403.5,99.67) .. (403.5,95) ;
\draw  [color={rgb, 255:red, 0; green, 0; blue, 255 }  ,draw opacity=1 ] (140,144) .. controls (140,148.67) and (142.33,151) .. (147,151) -- (165.03,151) .. controls (171.7,151) and (175.03,153.33) .. (175.03,158) .. controls (175.03,153.33) and (178.36,151) .. (185.03,151)(182.03,151) -- (201.5,151) .. controls (206.17,151) and (208.5,148.67) .. (208.5,144) ;
\draw  [color={rgb, 255:red, 0; green, 0; blue, 255 }  ,draw opacity=1 ] (209.5,144) .. controls (209.36,148.67) and (211.62,151.07) .. (216.29,151.2) -- (216.3,151.2) .. controls (222.97,151.4) and (226.23,153.83) .. (226.09,158.49) .. controls (226.23,153.83) and (229.63,151.6) .. (236.29,151.79)(233.29,151.7) -- (236.3,151.79) .. controls (240.96,151.92) and (243.36,149.66) .. (243.5,145) ;
\draw    (369.5,213) -- (405,213) ;
\draw [shift={(405,213)}, rotate = 180] [color={rgb, 255:red, 0; green, 0; blue, 0 }  ][line width=0.75]    (0,5.59) -- (0,-5.59)   ;
\draw [shift={(369.5,213)}, rotate = 180] [color={rgb, 255:red, 0; green, 0; blue, 0 }  ][line width=0.75]    (0,5.59) -- (0,-5.59)   ;
\draw  [dash pattern={on 0.84pt off 2.51pt}]  (297.5,213) -- (369.5,213) ;

\draw  [dash pattern={on 4.5pt off 4.5pt}]  (261.5,213) -- (297.5,213) ;
\draw [shift={(297.5,213)}, rotate = 180] [color={rgb, 255:red, 0; green, 0; blue, 0 }  ][line width=0.75]    (0,5.59) -- (0,-5.59)   ;
\draw [shift={(261.5,213)}, rotate = 180] [color={rgb, 255:red, 0; green, 0; blue, 0 }  ][line width=0.75]    (0,5.59) -- (0,-5.59)   ;
\draw  [dash pattern={on 4.5pt off 4.5pt}]  (139,213) -- (210.5,213) ;
\draw [shift={(210.5,213)}, rotate = 180] [color={rgb, 255:red, 0; green, 0; blue, 0 }  ][line width=0.75]    (0,5.59) -- (0,-5.59)   ;
\draw [shift={(140.5,213)}, rotate = 180] [color={rgb, 255:red, 0; green, 0; blue, 0 }  ][line width=0.75]    (0,5.59) -- (0,-5.59)   ;
\draw  [dash pattern={on 4.5pt off 4.5pt}]  (210.5,213) -- (243.5,213) ;
\draw [shift={(243.5,213)}, rotate = 180] [color={rgb, 255:red, 0; green, 0; blue, 0 }  ][line width=0.75]    (0,5.59) -- (0,-5.59)   ;
\draw [shift={(210.5,213)}, rotate = 180] [color={rgb, 255:red, 0; green, 0; blue, 0 }  ][line width=0.75]    (0,5.59) -- (0,-5.59)   ;
\draw  [dash pattern={on 4.5pt off 4.5pt}]  (243.5,213) -- (261.5,213) ;
\draw [shift={(261.5,213)}, rotate = 180] [color={rgb, 255:red, 0; green, 0; blue, 0 }  ][line width=0.75]    (0,5.59) -- (0,-5.59)   ;
\draw [shift={(243.5,213)}, rotate = 180] [color={rgb, 255:red, 0; green, 0; blue, 0 }  ][line width=0.75]    (0,5.59) -- (0,-5.59)   ;


\draw (90,86) node [scale=0.8]  {$\beta _{1} =\upsilon _{1}$};
\draw (178,117) node [scale=0.8]  {$\upsilon _{1}$};
\draw (279,64) node [scale=0.8]  {$1$};
\draw (229,163) node [scale=0.8]  {$\upsilon _{2}( 1-\upsilon _{1})$};
\draw (90,140) node [scale=0.8]  {$\beta _{2} =\upsilon _{2}( 1-\upsilon _{1})$};
\draw (90, 180) node [scale = 0.8] {$ \vdots $};
\draw (90, 215) node [scale = 0.8] {$\beta_{\infty} $};
\draw (177,163) node [scale=0.8]  {$\upsilon _{1}$};
\draw (311,118) node [scale=0.8]  {$1-\upsilon _{1}$};
\draw (175,225) node [scale=0.8]  {$\beta _{1}$};
\draw (228,225) node [scale=0.8]  {$\beta _{2}$};
\draw (254,225) node [scale=0.8]  {$\beta _{3}$};
\draw (278,225) node [scale=0.8]  {$\beta _{4}$};
\draw (316,225) node [scale=0.8]  {$\beta _{i}$};
\draw (387,225) node [scale=0.8]  {$\beta _{\infty }$};

\end{tikzpicture}}
\caption{Illustration of the stick-breaking model to construct $\beta$ for GEM($\gamma$). The probabilities of $\beta_{k} $ are given by a procedure resembling the breaking of a unit-length stick $\beta_{k} = \upsilon_{k}\prod_{i=1}^{k-1}(1-\upsilon_{i})$ with $v_{k}$ independently drawn from a Beta distribution, $\mathrm{Beta}(1, \gamma)$.}
\label{fig:stickbreaking}
\end{figure}

To construct a valid probability measure $\pi_{z_{t-1}}$ (in (\ref{eq:latentpattern})) that can describe the transition probability while automatically increasing the number of interaction patterns and synchronously assigning new pattern parameters $\theta_{z_{t}}$ (in (\ref{eq:emission})) with new data get observed incrementally, we introduce the Dirichlet process (DP) over an infinite measure space as the basis. The DP can be constructed by two independent valid probability measures:  
\begin{equation}
G \sim \sum_{k=1}^{\infty} \beta_{k} \delta_{\theta_{k}}
\end{equation}
where $\beta = [\beta_{1}, \beta_{2}, \dots]$ are drawn from the $\mathrm{GEM}(\gamma)$ distribution\footnote[2]{The Griffiths-Engen-McCloskey (GEM) distribution is a special case of DP, which can be reconstructed via the stick-breaking model.} with  the constraint $\sum_{i=1}^\infty \beta_i=1$,  denoted as $\beta \sim {\mathrm{GEM}}(\gamma)$ (see Fig. \ref{fig:stickbreaking}), and $\theta_{k}$ is drawn from a base measure  $H$. The $\delta_{\theta_{k}}$ is the Dirac delta measure and denotes a point mass at $\theta_{k}$. Note that the base measure  $H$ can be continuous or discrete, but $G$ is a discrete probability measure with the impact of the Dirac delta measure. Moreover, both $\beta_{k}$ and $\theta_{k}$ are random, thus indicating that $G$ becomes a random probability measure and can be rewritten in the form of DP, $G\sim \mathrm{DP}(\gamma, H)$. The above discussion implies that $\gamma$, called concentration parameter, governs how close the random probability measure $G$ is to the base measure $H$. 

\begin{figure}[t]
\centering
\resizebox{0.97\linewidth}{!}{\begin{tikzpicture}[yscale=1,xscale=1]
\draw [-] (-4,0) -- (4,0);
\draw [samples=100, domain=-3:3, blue] plot (\x, {3*exp(-1*pow((\x),2))});

\draw (0,0) -- (0,1);
\filldraw (0,1) circle(1.5pt);

\draw (-1,0) -- (-1,0.4);
\filldraw (-1,0.4) circle(1.5pt);
\draw (-1, -0.25) node [font=\small] {$\theta_{1}$};

\draw (-2,0) -- (-2,0.2);
\filldraw (-2,0.2) circle(1.5pt);
\draw (-2, -0.25) node [font=\small] {$\theta_{2}$};

\draw (-2.6,0) -- (-2.6,0.6);
\filldraw (-2.6,0.6) circle(1.5pt);

\draw (2.5,0) -- (2.5,1);
\filldraw (2.5,1) circle(1.5pt);
\draw (2.5, -0.25) node [font=\small] {$\theta_{3}$};

\draw(-0.2,0) -- (-0.2,0.3);
\filldraw (-0.2,0.3) circle(1.5pt);

\draw(0.05,0) -- (0.05,0.45);
\filldraw (0.05,0.45) circle(1.5pt);

\draw [red] (-0.5,0) -- (-0.5,1.75);
\filldraw [red] (-0.5,1.75) circle(1.5pt);
\draw [decorate, decoration={brace,amplitude=5pt}, xshift=-1pt, yshift=0pt, red](-0.55,0) -- (-0.55,1.75) node [red,midway,xshift=-0.33cm]
{\footnotesize $\beta_k$};
\draw (-0.5, -0.25) node [font=\small, red] {$\theta_{k}$};

\draw(0.15,0) -- (0.15,0.4);
\filldraw (0.15,0.4) circle(1.5pt);

\draw(0.23,0) -- (0.23,0.6);
\filldraw (0.23,0.6) circle(1.5pt);

\draw(0.73,0) -- (0.73,0.7);
\filldraw (0.73,0.7) circle(1.5pt);

\draw(1.13,0) -- (1.13,0.45);
\filldraw (1.13,0.45) circle(1.5pt);
\draw (1.25, -0.25) node [font=\small] {$\theta_{\infty}$};

\draw [-] (-1,0) -- (-1,0.4);

\draw (-2.5,2) node {$G_{0}\sim \sum_{k=1}^{\infty} \beta_{k}\delta_{\theta_{k}}$};
\draw (-2.5,1.45) node {or};
\draw (-2.5,1) node {$G_{0}\sim \mathrm{DP}(\gamma,H)$};
\draw [blue] (1,2) node {$H$};

\draw (0.3, -0.25) node [font=\small] {$\dots$};
\end{tikzpicture}}
\resizebox{0.97\linewidth}{!}{\begin{tikzpicture}[yscale=1,xscale=1]
\draw [-] (-4,0) -- (4,0);

\draw [blue] (0,0) -- (0,1);
\draw (-0.01,0) -- (-0.01,0.1);
\filldraw (-0.01,0.1) circle(1.5pt);

\draw [blue](-1,0) -- (-1,0.4);
\draw (-1.01, -0.25) node [font=\small] {$\theta_{1}$};
\draw (-1.01,0) -- (-1,0.1);
\filldraw (-1.01,0.1) circle(1.5pt);

\draw [blue](-2,0) -- (-2,0.2);
\draw (-2, -0.25) node [font=\small] {$\theta_{2}$};
\draw (-2.01,0) -- (-2.01,0.65);
\filldraw (-2.01,0.65) circle(1.5pt);

\draw [blue](-2.6,0) -- (-2.6,0.6);
\draw (-2.61,0) -- (-2.61,0.2);
\filldraw (-2.61,0.2) circle(1.5pt);

\draw [blue](2.5,0) -- (2.5,1);
\draw (2.5, -0.25) node [font=\small] {$\theta_{3}$};
\draw (2.49,0) -- (2.49,0.3);
\filldraw (2.49,0.3) circle(1.5pt);

\draw [blue](-0.2,0) -- (-0.2,0.3);
\draw (-0.21,0) -- (-0.21,0.25);
\filldraw (-0.21,0.25) circle(1.5pt);

\draw [blue](0.05,0) -- (0.05,0.45);
\draw (0.04,0) -- (0.04,0.55);
\filldraw (0.04,0.55) circle(1.5pt);

\draw [blue] (-0.5,0) -- (-0.5,1.75);
\draw [decorate, decoration={brace,amplitude=5pt}, xshift=-1pt, yshift=0pt, red](-0.55,0) -- (-0.55,1.50) node [red,midway,xshift=-0.4cm]
{\footnotesize $\pi_{ik}$};
\draw (-0.49,0) -- (-0.49,1.50);
\filldraw (-0.49,1.50) circle(1.5pt);

\draw (-0.5, -0.25) node [font=\small, red] {$\theta_{k}$};

\draw [blue](0.15,0) -- (0.15,0.4);
\draw (0.14,0) -- (0.14,0.2);
\filldraw (0.14,0.2) circle(1.5pt);

\draw [blue](0.23,0) -- (0.23,0.6);
\draw (0.22,0) -- (0.22,0.5);
\filldraw (0.22,0.5) circle(1.5pt);

\draw [blue](0.73,0) -- (0.73,0.7);
\draw (0.72,0) -- (0.72,0.1);
\filldraw (0.72,0.1) circle(1.5pt);

\draw [blue](1.13,0) -- (1.13,0.45);
\draw (1.12,0) -- (1.12,0.56);
\filldraw (1.12,0.56) circle(1.5pt);
\draw (1.25, -0.25) node [font=\small] {$\theta_{\infty}$};

\draw [blue][-] (-1,0) -- (-1,0.4);

\draw (-2.5,1.4) node {$G_{i}\sim \mathrm{DP}(\alpha, G_{0})$};
\draw (-2.5,0.9) node {$i=1,2,\cdots$};
\draw [blue] (1,1.2) node {$G_0$};

\draw (0.3, -0.25) node [font=\small] {$\dots$};
\end{tikzpicture}}
\caption{Illustration of DP construction with a continuous base measure $H$ and HDP construction with a discrete base measure $G_{0}$ drawn from the previous DP, where $G_i$ is one instance drawn from $G_{0}$. The sum of the height of all the vertical lines is equal to one when $k\rightarrow \infty$.}
\label{fig_dp_hdp}
\end{figure}
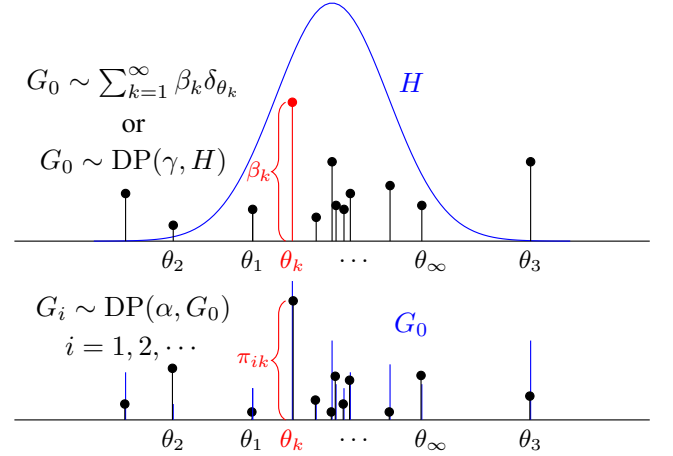


 When $H$ is continuous, a group of discrete distributions $G$ drawn from  $\mathrm{DP}(\gamma, H)$ directly with a specific base measure  $H$ would not share atoms across different $G$; that is, the probability of drawing two  samples with the same value from a continuous space is zero. In order to make the atoms $\{\theta_{k}\}$ among $G$ shareable, we add another DP layer to discretize the continuous base measure  $H$, thus reshaping a hierarchical Dirichlet process (HDP)\cite{teh2005sharing}, i.e., $G_{0} \sim \mathrm{DP}(\gamma, H)$, $G \sim \mathrm{DP}(\alpha, G_{0})$, as illustrated in Fig. \ref{fig_dp_hdp}.

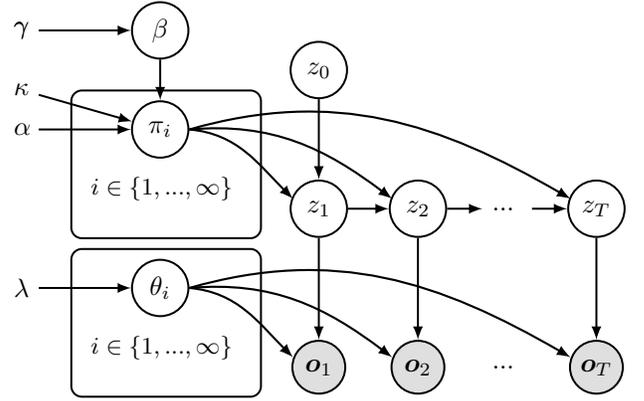
\begin{figure}[t]
\centering
\tikzset{every picture/.style={line width=0.75pt}} 

\begin{tikzpicture}[x=0.75pt,y=0.75pt,yscale=-1,xscale=1]
\node [obs] (o1) at (20,20) {$\boldsymbol{o}_1$};
\node [circle,draw=black,fill=white, inner sep=0pt,minimum size=0.75cm] (p1) at (20,-60) {$z_1$};
\node [circle,draw=black,fill=white, inner sep=0pt,minimum size=0.75cm] (p0) at (20,-130) {$z_0$};
\node [obs] (o2) at (70,20) {$\boldsymbol{o}_2$};
\node [circle,draw=black,fill=white, inner sep=0pt,minimum size=0.75cm] (p2) at (70,-60) {$z_2$};
\node [obs] (o3) at (160,20) {$\boldsymbol{o}_T$};
\node [circle,draw=black,fill=white, inner sep=0pt,minimum size=0.75cm] (p3) at (160,-60) {$z_T$};
\node [circle,draw=black,fill=white,inner sep=0pt,minimum size=0.75cm] (theta) at (-60,-20) {$\theta_i$};
\node [circle,draw=black,fill=white,inner sep=0pt,minimum size=0.75cm] (pi) at (-60,-100) {$\pi_i$};
\node [circle,draw=black,fill=white,inner sep=0pt,minimum size=0.75cm] (beta) at (-60,-150) {$\beta$};
\node [text width=2cm] (inf1) at (-57,10) {\small{$i\in\left\{1,...,\infty \right\}$}};

\node [text width=2cm] (inf2) at (-57,-70)
{\small{$i\in\left\{1,...,\infty \right\}$}};

\node [text width=.2cm] (lambda) at (-130,-20) {$\lambda$};
\node [text width=.2cm] (alpha) at (-130,-100) {$\alpha$};
\node [text width=.2cm] (kappa) at (-130,-120) {$\kappa$};
\node [text width=.2cm] (gamma) at (-130,-150) {$\gamma$};
\node [text width=.2cm] (gamma) at (-130,-150) {$\gamma$};
\node [text width=.4cm] (1dot) at (115,-60) {$...$};
\node [text width=.4cm] (2dot) at (115,20) {$...$};
\path [draw,->] (p2) edge (1dot);
\path [draw,->] (1dot) edge (p3);
\path [draw,->] (p1) edge (o1);
\path [draw,->] (p2) edge (o2);
\path [draw,->] (p3) edge (o3);
\path [draw,->] (p1) edge (p2);
\path [draw,->] (p0) edge (p1);
\path [draw,->] (alpha) edge (pi);
\path [draw,->] (kappa) edge (pi);
\path [draw,->] (beta) edge (pi);
\path [draw,->] (gamma) edge (beta);
\path [draw,->] (lambda) edge (theta);
%
\plate [color=black] {part2} {(theta)(inf1)} { };
\plate [color=black] {part3} {(pi)(inf2)} { };
\draw[black,->,thick] (pi.east) to [in=-150,out=-16] (147,-65);
\draw[black,->,thick] (pi.east) to [in=-143,out=-5] (57,-65);
\draw[black,->,thick] (pi.east) to [in=-135,out=5] (7,-65);
\draw[black,->,thick] (theta.east) to [in=-150,out=-16] (147,15);
\draw[black,->,thick] (theta.east) to [in=-143,out=-5] (57,15);
\draw[black,->,thick] (theta.east) to [in=-135,out=5] (7,15);

\end{tikzpicture}
\caption{The graphical model of the sticky HDP-HMM. Gray-filled circles represent the learned representations, the white circles are the random variables, and the box denotes a set of infinite collections of random variables.}
\label{fig5}
\end{figure}

\subsubsection{Sticky HDP-HMM}
In HDP, we expect that the transition of interaction patterns between two adjacent data frames is adjustable and has a low frequency. Therefore, an extra sticky parameter $\kappa>0$ is added to bias the process toward self-transition in the HDP, called the sticky HDP. Thus, placing a sticky HDP prior over infinite transition matrices of HMM, we obtain a sticky HDP-HMM$(\kappa, \alpha, H) $\cite{fox2011sticky} (as shown in Fig. \ref{fig5}), formulated as
\begin{subequations}
\begin{align}
\beta|\gamma &\sim {\mathrm{GEM}}(\gamma)\label{eq14a}\\
\theta_i|\lambda &\mathop{\sim}^{iid} H(\lambda),\ i=1,2,\cdots\label{eq14b}\\
\pi_i|\alpha, \beta, \kappa &\mathop{\sim}^{iid} \mathrm{DP}(\alpha+\kappa,\frac{\alpha \beta + \kappa \delta_i}{\alpha+\kappa}),\ i=1,2,\cdots\label{eq14c}\\
z_t|z_{t-1}&\sim\pi_{z_{t-1}},\ t=1,2,\cdots,T\label{eq14d}\\
\boldsymbol{o}_t|z_t,\theta_{z_t} &\sim F(\theta_{z_t}),\ t=1,2,\cdots,T\label{eq14e}
\end{align}
\end{subequations}
where (\ref{eq14a}) and (\ref{eq14b}) are the DP shared global priors (the first DP), (\ref{eq14c}) is the transition matrix prior (the second DP), (\ref{eq14d}) generates the latent states, and (\ref{eq14e}) draws the observations.

Once built the generative model, we  perform the approximate inferences of the model parameters  using the sequential  data, i.e., sequential representatives of multi-vehicle interactions.  We implemented a weak-limit Gibbs sampling algorithm for the sticky HDP-HMM\cite{johnson2013hdphsmm}. The observation function $F(\theta_{z_{t}})$ is treated as a Gaussian distribution specified by $\theta_{z_{t}}=[\mu_{z_{t}},\Sigma_{z_{t}}]$, we take $\mu_{z_{t}}=0$ according to \cite{hamada2016modeling}  and the hyperparameters for $\theta$ are assumed to share an Inverse-Wishart prior\cite{wang2018driving}.  We set  a beta prior on $\kappa/(\alpha+\kappa)$, and place vague gamma priors on the hyperparameters $\gamma$ and $(\alpha+\kappa)$ in order to make the posterior distribution computationally tractable.

\section{Data Processing and Experimental Result Analysis}\label{sec5}

\subsection{Data Preprocessing}

\begin{figure}[t]
\centering
\includegraphics[width = \linewidth]{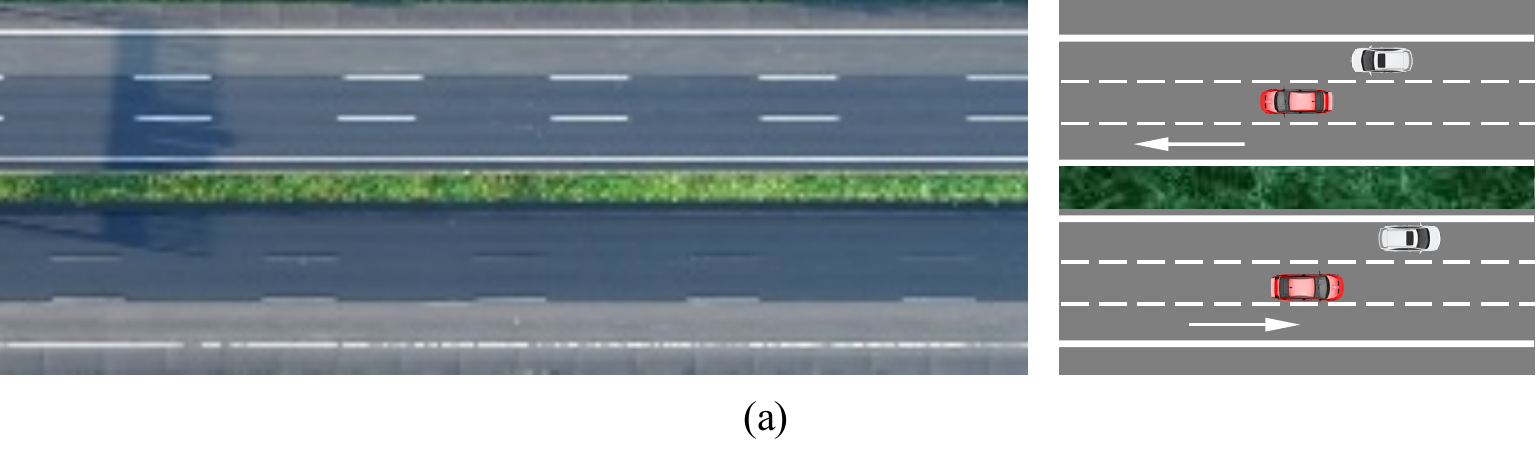}\\
\includegraphics[width = \linewidth]{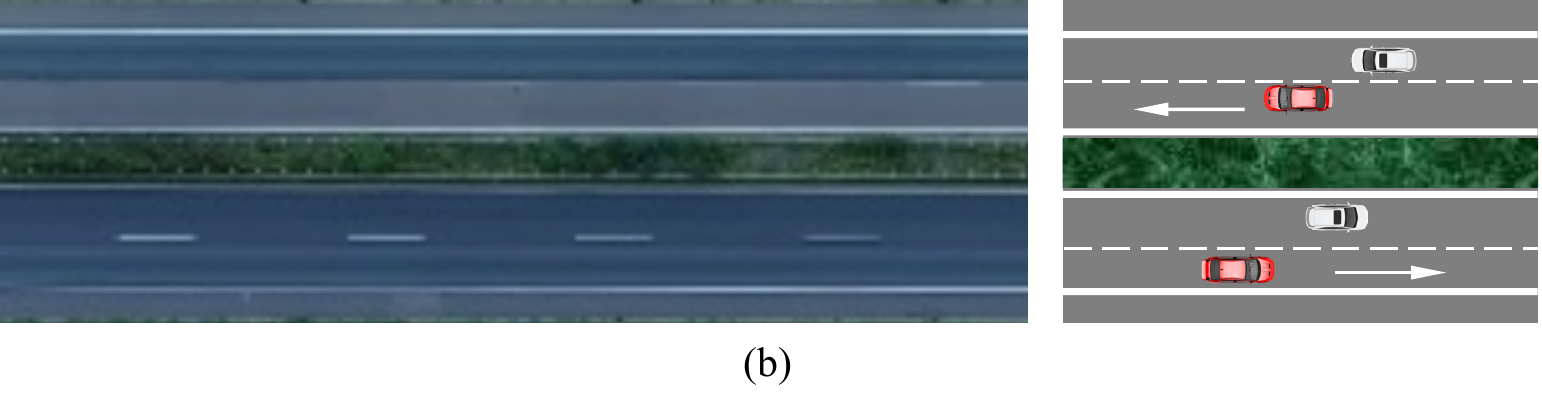}\\
\caption{Illustrations of (a) the six-lane divided highway and (b) the four-lane divided highway in the highD dataset.}
\label{fig:highway}
\end{figure}

We trained and validated our proposed framework using the lane-change data from the released Highway Drone (highD) Dataset\cite{highDdataset}. The highD dataset has $60$ video recordings, logged with the sampling frequency of $25$ Hz, consisting of two different environment settings (see Fig. \ref{fig:highway}): four-lane divided and six-lane divided. We selected one record for each environment setting -- Recording$03$ for the four-lane divided highway and Recording$07$ for the six-lane divided highway, as shown in Table \ref{table1}. To reduce the required computer memory, we downsampled the raw sequential frames to $5$ Hz, thus obtaining $14563$ frames of all lane-change vehicles.


Furthermore, the $y$-axis in the highD dataset is heading down and different from what we defined in Fig. \ref{fig:lanechangescenario}, we transformed the original coordinate system by rotating the $y$-axis in $180$ deg clock-wisely. Meanwhile, we also rotated the coordinates of the heading-left vehicles in $180$ deg clock-wisely. This pre-processing unifies the move directions of all vehicles (i.e., heading right), and the coordinates in some results could be found with negative values.

\begin{table}[t]
    \caption{Metadata for the highD Dataset.}
    \begin{center}
    \label{table1}
    \begin{tabular}{c c c c}
        \hline
        \hline
        { }& {Recording03} & {Recording07} & {Total}\\
        \hline
        Total vehicles & $914$ & $855$ & $1769$\\
        Lane-change vehicles & $119$ & $130$ & $249$\\
        Total frames & $303488$ & $257031$ & $560519$\\
        Lane-change frames* & $35333$ & $37270$ & $72603$\\
        \hline\hline
        \multicolumn{4}{l}{* All the frames of lane-change vehicles.}
    \end{tabular}
    \end{center}
\end{table}

\subsection{AS-GVF Feature Representation}

\begin{figure}[t]
\centering
\includegraphics[width = \linewidth]{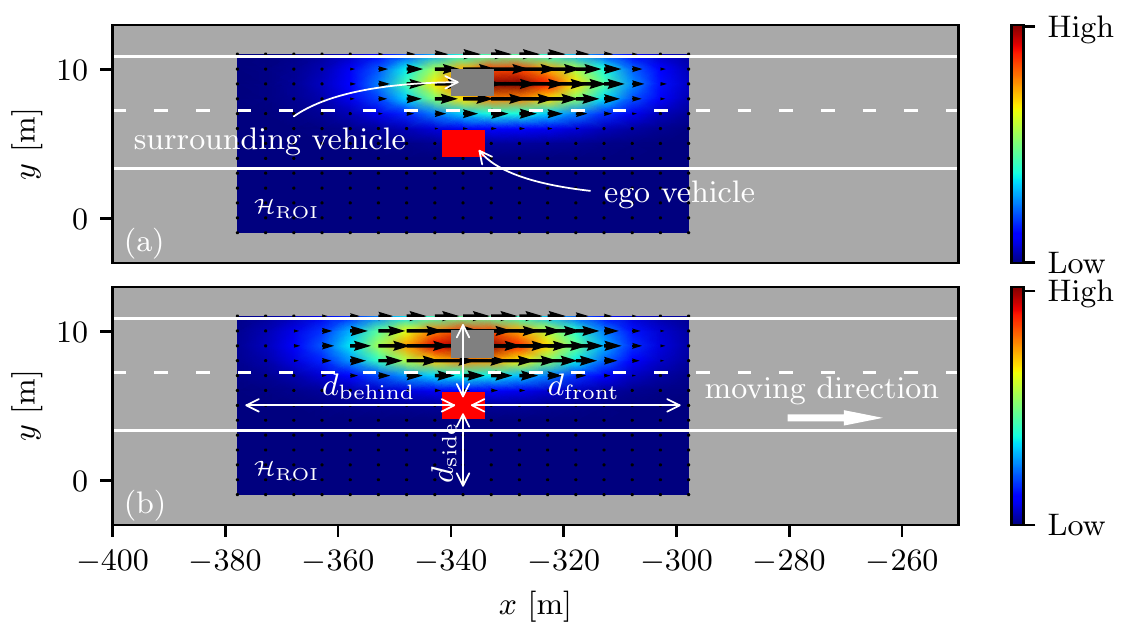}
\caption{ Comparison of (a) our AS-GVF considering acceleration sensitivity with (b) the conventional GVF.}
\label{fig2}
\end{figure}


In AS-GVF, we define $\mathcal{H}_{\mathrm{ROI}}$ as a rectangle area symmetrically centered on the ego vehicle and specified by three distances to the center of the ego vehicle (see Fig. \ref{fig2}): the front distance, $d_{\mathrm{front}}$, the behind distance, $d_{\mathrm{behind}}$, and the left/right distances $d_{\mathrm{side}}$. The lane width is around $4$ m in the environment; therefore, we set the left/right distance as $d_{\mathrm{side}}=6$ m to cover the left and right lanes. For the longitudinal direction, we select a medium distance according to \cite{yan2015classifying, xu2017ego} and set $d_{\mathrm{front}}=d_{\mathrm{behind}}=40$ m. Theoretically, the defined ROI could be infinitely large, and the length and width of the ROI can be meshed into as many grids as expected. Considering the computational feasibility, the AS-GVF is constructed over the grid points in the ROI by meshing the width and length with intervals of $1$ m and $5$ m. Thus, the numbers of testing locations in $\mathcal{H}_{\mathrm{ROI}}$ for our AS-GVF along the $x$ and $y$ directions are $17$ and $13$. In this way, a tensor with a size of $13\times17\times2$ describes the AS-GVF of each frame, where $2$ represents the velocity components in the $x$ and $y$ directions. In the AS-GVF model, we set $A=1$, $\sigma_x=15$ m, $\sigma_y=1.5$ m, $\lambda_x=0.6$, and $\lambda_y=0.9$.  Note that this setting is just tuned based on experience, their values should be adapted to set up carefully for other applications.

Figure \ref{fig2} compares our developed AS-GVF with the GVF regarding the capability of representing multi-vehicle interactions, in which the ego vehicle is heading to the right while the surrounding vehicle is accelerating and trying to overtake. 
Black arrows represent the direction and size of the estimated relative velocity ($\Delta v_x$, $\Delta v_y$) at any location in $\mathcal{H}_{\mathrm{ROI}}$ according to the observed relative speed of these two vehicles and their states. 
Comparison verifies that the conventional GVF can only capture the symmetric influence of the surrounding vehicle on its surroundings but ignore the vehicle's acceleration, as shown by the symmetric heatmap in Fig. \ref{fig2}(b). However, our developed AS-GVF can account for the influence of acceleration on its nearby region of the surrounding vehicle, as shown in Fig. \ref{fig2}(a). The surrounding vehicle has a higher density (i.e., a stronger influence) at its front area than its behind due to the accelerating maneuvers. In summary, our developed AS-GVF is flexible to adapt to the driving intent via the incorporation of acceleration/deceleration.


\begin{figure}[t]
    \centering
    \includegraphics[width=\linewidth]{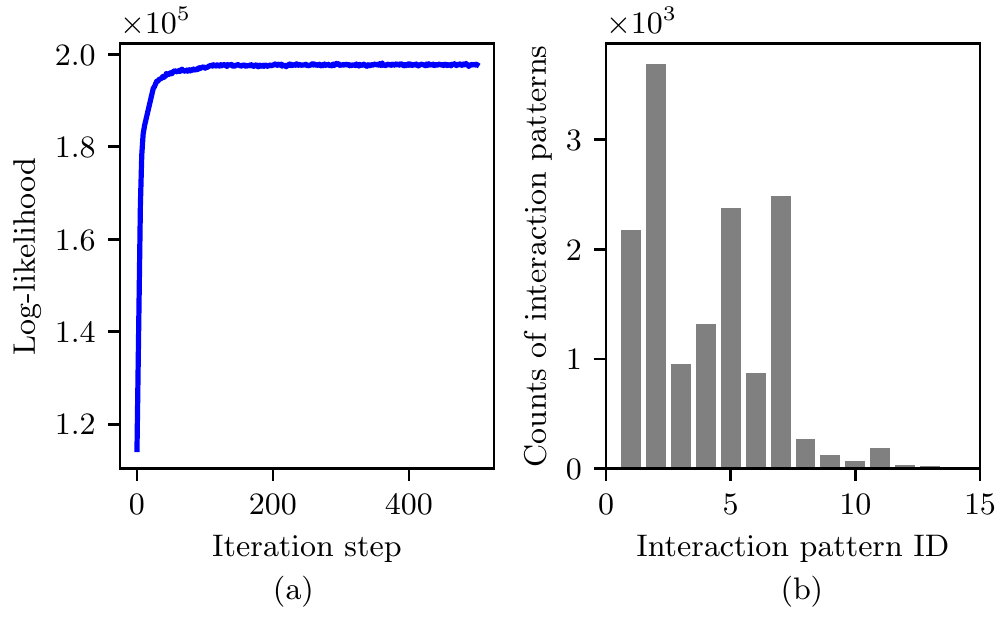}
    \caption{(a) The log-likelihood of training data with respect to the iterations and (b) the histogram of the number of interaction patterns during lane change.}
    \label{fig7}
\end{figure}
\begin{figure}[t]
\centering
\includegraphics[width = \linewidth]{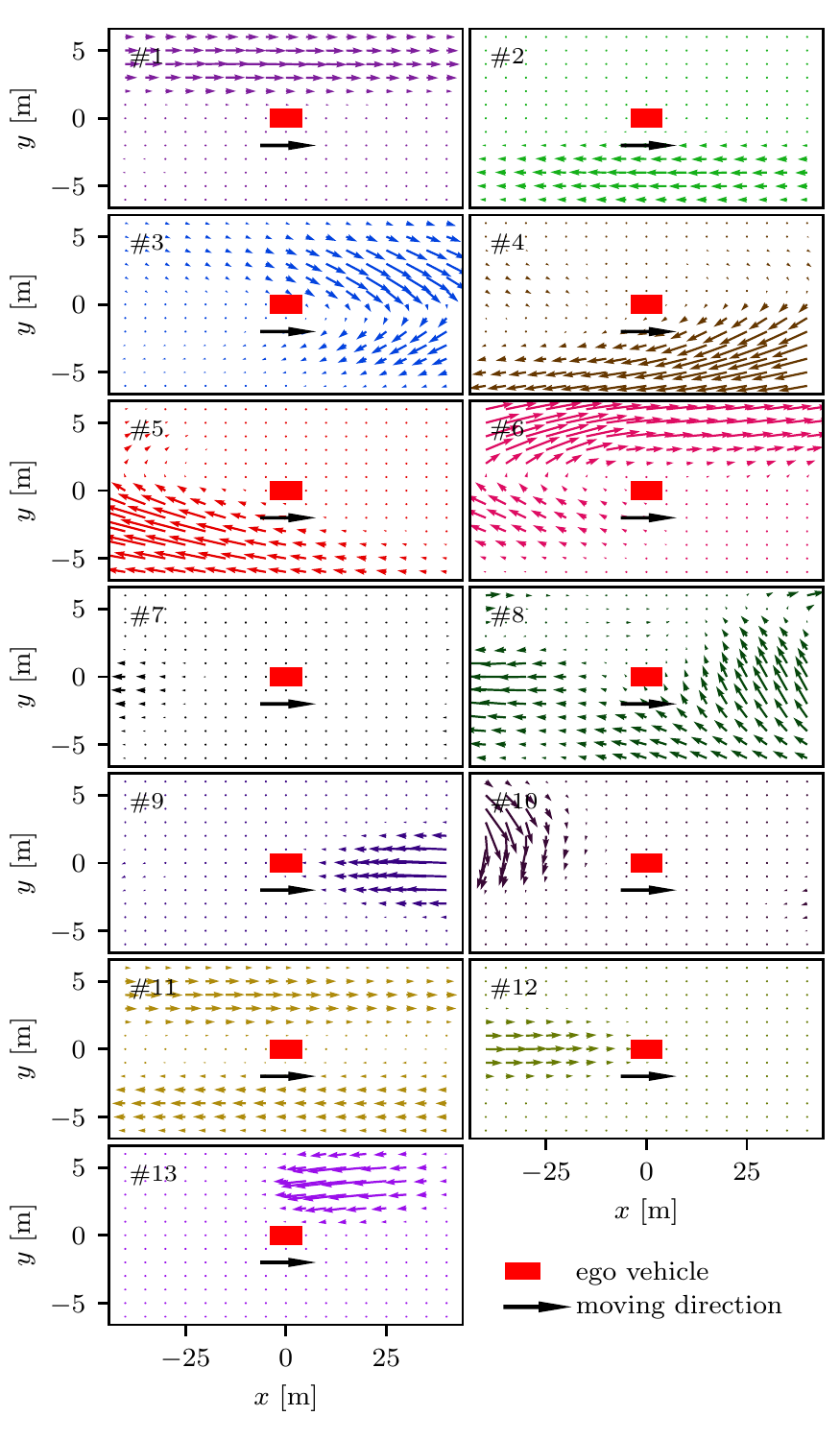}
\caption{The representative fields of the learned $13$ patterns. The red-shaded rectangles represent the ego vehicle which is moving toward the right direction.}
\label{fig_result}
\end{figure}

\subsection{Convolutional Autoencoder}\label{sec_CAE}

Figure \ref{fig_CAE} details the CAE architecture, constructed with an asymmetric structure by adding an extra transpose convolutional layer before the output layer. We used the Nesterov-accelerated Adaptive Moment Estimation (Nadam) algorithm \cite{dozat2016incorporating} to optimize our model with a mini-batch size of $1024$ on NVIDIA TESLA P$40$. All of the $14563$ frames are set as training data. We used the mean squared error (MSE) as the reconstruction loss function, which is reduced to as low as $8.77\times10^{-5}$ after $2\times10^4$ iterations.

In order to built a representative feature that can capture both the interaction of the ego vehicle with their surroundings and their driving behavior at each frame, we use the combination (denoted as $\boldsymbol{o}$) of the ego vehicle state $[v_x, v_y, a_x, a_y ]^{\top} \in \mathbb{R}^{4\times 1}$ and the extracted latent codes $h\in \mathbb{R}^{8\times 1}$ via CAE. Thus, the combination over all frames forms a sequential representation $O = \{\boldsymbol{o}_{1}, \dots, \boldsymbol{o}_{T}  \}$ with a size of $12 \times T$, where $\boldsymbol{o}_{t} \in \mathbb{R}^{12\times 1}$ is the synthesized feature of interactions at time $t$, and $T$ is the total number of after-processing frames, $T = 14563$.

\subsection{Interaction Pattern Learning and Spatial Analysis}
In order to extract patterns from sequential lane-change interactions, the above extracted sequential features $O$ are treated as the training data and passed through the sticky HDP-HMM\cite{johnson2013hdphsmm} with the maximum training iterations of $500$. Fig. \ref{fig7}(a) shows the log-likelihood of the data with parameter updates along with the training iterations. It indicates that the training procedure is convergent. Fig. \ref{fig7}(b) shows the total $13$ types of interaction patterns during lane change, indicating that during lane-change scenarios, human drivers interact with their surrounding vehicles by following patterns \#$1$ $\sim$ \#$7$ more frequently than following patterns \#$8$ $\sim$ \#$13$.

\begin{figure}[t]
\centering
\includegraphics[width = \linewidth]{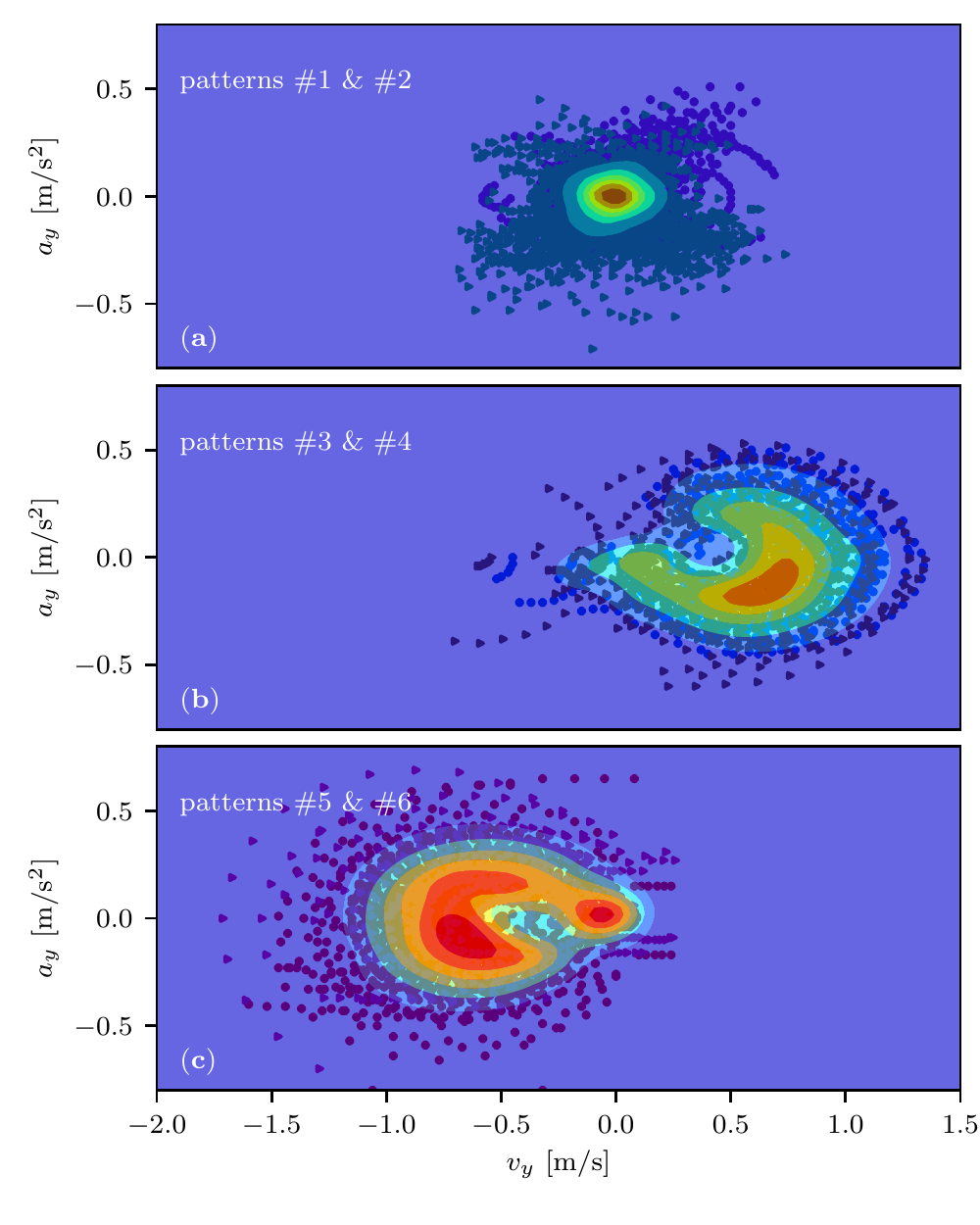}
\caption{The distribution of lateral operation states for six typical patterns $\#1-\#6$ in the $v_y - a_{y}$ coordinates.}%
\label{fig_decision}
\end{figure}

\begin{figure*}[t]
\centering
\includegraphics[width = \linewidth]{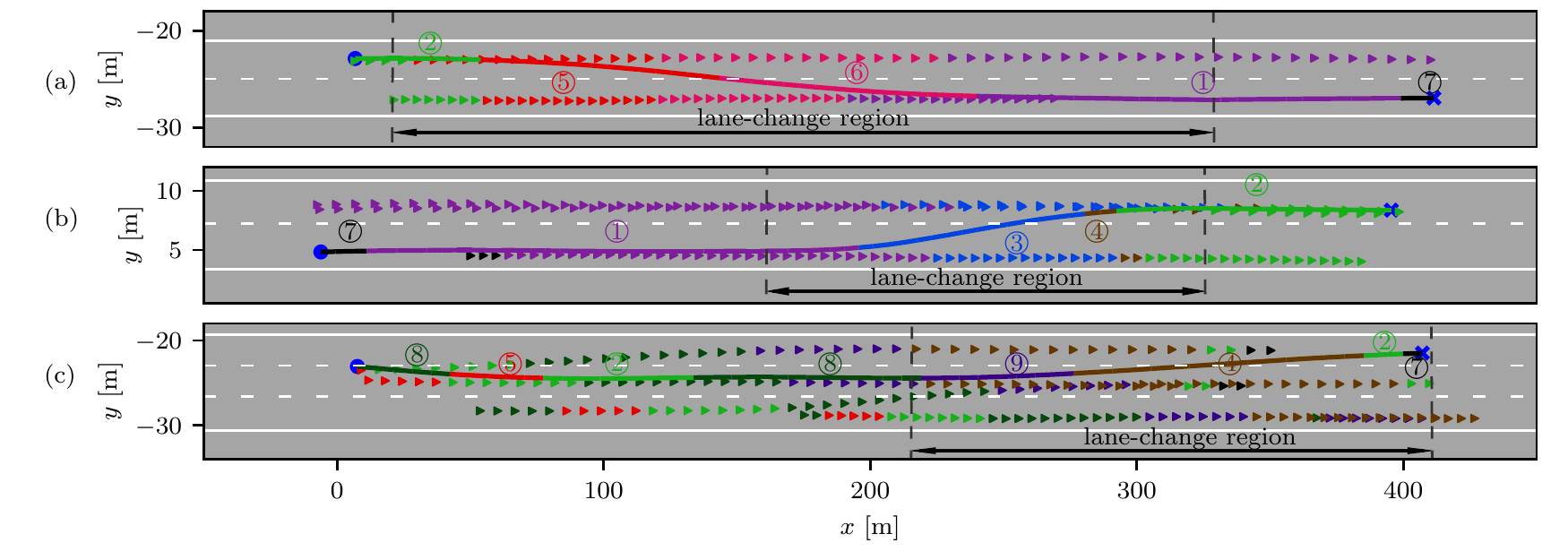}
\caption{Three examples of illustrating the dynamic process over interaction patterns during lane change in different road contextual settings: (a), (b) four-lane divided highway and (c) six-lane divided highway. While solid and dashed line -- Lane marker; Circled number -- Pattern ID; Colored solid line -- Trajectory of the ego vehicle; Filled triangle ($\blacktriangleright$) -- Trajectory of surrounding vehicles; Blue dot ({\color{blue} $\bullet$}) -- Start position of the ego vehicle; and Blue cross ({\color{blue} $\boldsymbol{\times}$}) -- End position of the ego vehicle.}
\label{demo_dynamic}
\end{figure*}

Figure \ref{fig_result} lists all types of interaction patterns with velocity fields distinguished by colors, which enables the interpretation of these patterns on how they characterize the ego vehicle's interactions with their surrounding vehicles. Each subplot is computed by averaging all of the velocity fields with the same pattern label. The colored arrows represent the estimated relative velocity at any location in $\mathcal{H}_{\mathrm{ROI}}$, and a long arrow indicates a high relative speed with respect to the ego vehicle at that location. In what follows, we will semantically interpret each interaction pattern.

Pattern \#$1$ (Pattern \#$2$) describes that a surrounding vehicle is moving faster (slower) in the adjacent left-hand (right-hand) lane, thus causing a relative velocity pointing in the head-forward (backward) direction. The opposite sides of both patterns -- the adjacent right-hand lane in pattern \#$1$ and the adjacent left-hand lane in pattern \#$2$ -- possess near-zero arrows, indicating that there are no vehicles on associated lanes, or that some vehicles are on the associated lanes but with a near-zero relative velocity to the ego vehicle. In these two interaction patterns, the human driver keeps moving on the current driving lane straightly; therefore, it leads to an imperceptible tendency of lateral moving. Correspondingly, Fig.\ref{fig_decision}(a) shows the distribution of the lateral moving states $[v_{y}, a_{y}]$ of the ego vehicle and illustrates that all states are almost falling around (0,0) -- both the lateral speed and acceleration are near-zero. Further, these two interaction patterns typically occur in the lane-change preparation stage, as shown in Fig. \ref{demo_dynamic}. One interesting finding is that pattern \#$11$ is likely the combination of patterns \#$1$ and \#$2$. It only occurs on the six-lane divided highway (see Fig. \ref{fig:highway}(a)) because it depicts the interaction scenario in which the ego vehicle is moving faster than the vehicles on its adjacent right-hand lane while slower than the vehicles on its adjacent left-hand lane.

Pattern \#$3$ represents the scenario that the ego vehicle starts accelerating and operating the vehicle to approach to the adjacent left-hand lane -- giving up following the slower leading vehicle on the current lane -- as the head of arrows at the left front area of the ego vehicle is pointing in the right front direction. Fig. \ref{demo_dynamic}(b) shows an associated real-world case for pattern \#$3$.  Pattern \#$4$ represents that the ego vehicle is changing lanes to the left, as shown in Fig. \ref{demo_dynamic}(b) and (c). Unlike pattern \#$3$ in which the right side area of the ego vehicle possesses near-zero arrows, pattern \#$4$ possesses near-zero arrows at the left side area of the ego vehicle, indicating that no vehicles exist on the left-hand target lane. Correspondingly, Fig. \ref{fig_decision}(b) shows the states of the ego vehicle regarding patterns \#$3$ and \#$4$, which implies that most left lane-change cases are specified by a positive $v_y$ and a negative $a_y$.

Pattern \#$5$ describes that the ego vehicle moves faster and passes a slower vehicle on its right side during lane change, while no vehicles exist in its left front area.  In pattern \#$6$, the ego vehicle is almost in the target lane and overtaking a slow vehicle; moreover, it represents the behavior of giving ways to other faster vehicles on the left lane, as shown in Fig. \ref{demo_dynamic}(a). The distribution of the ego vehicle's states corresponding to patterns \#$5$ and \#$6$ is given in Fig. \ref{fig_decision}(c), which implies that the ego vehicle slightly moves in the right direction during lane change, accompanying a negative lateral speed $v_y$.

Pattern \#$7$ describes a special interaction case, in which there is no obvious relative movement in the ROI because the length of arrows is close to zero. This interaction pattern typically occurs at the beginning or end of the lane-change procedure, as shown in Fig. \ref{demo_dynamic}. This explains why pattern \#$7$ has a large proportion of all interaction patterns, as displayed in Fig. \ref{fig7}(b). 

Pattern \#$8$ represents two cases: 1) the ego vehicle is cutting into the adjacent right-hand lane, and 2) the right leading vehicle is cutting into the front area of the ego vehicle. Thus, the arrow behind the ego vehicle is pointing in the opposite moving direction -- that is, the behind following vehicle decelerates to guarantee an appropriate safety gap -- while the arrows in the front right side of the ego vehicle are pointing in the left-behind direction. Correspondingly, pattern \#$10$ describes the opposite view of the right leading vehicle in pattern \#$8$. These cases typically occur in the extreme cluttered scenarios with a mess of surrounding vehicles,  as shown in Fig. \ref{demo_dynamic}(c).

\begin{figure*}[t]
    \centering 
    \subfloat{%
        \label{transistion_ALL_self}
        \includegraphics[width=\textwidth]{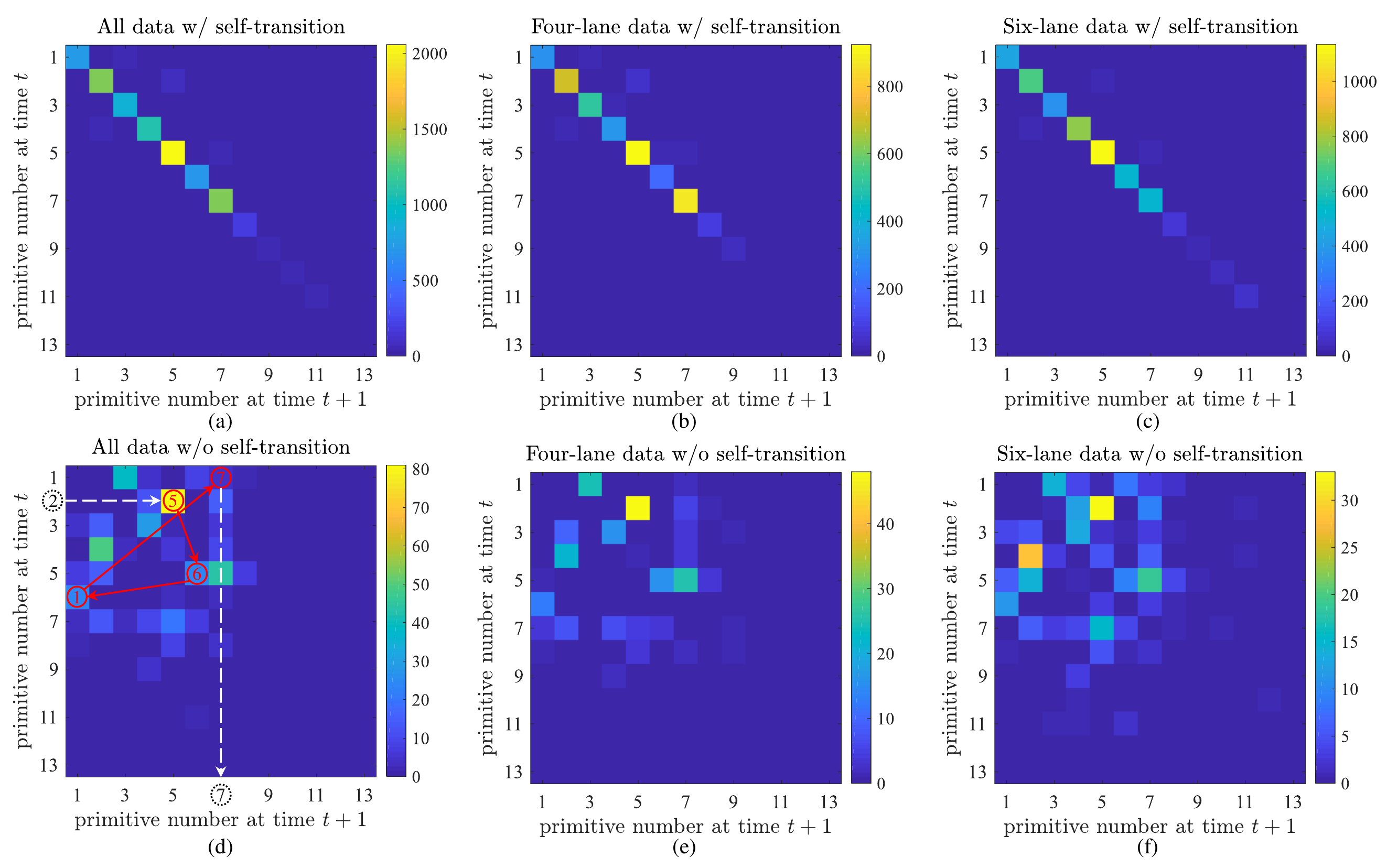}}%
    \caption{Transition frequency matrix of the extracted interaction patterns in lane-change region.}%
    \label{fig_transistion}
\end{figure*}

Both patterns \#$9$ and \#$12$ represent that the ego vehicle and the surrounding vehicle are approaching each other in the same lane. However, in pattern \#$9$, the ego vehicle is getting closer to the rear side of the ahead vehicle; the behind following vehicle in pattern \#$12$ is chasing the ego vehicle in the same direction. Pattern \#$13$ describes that the ego vehicle on the current lane drives faster than the surrounding vehicle on its adjacent left-hand lane, which is also a typical scenario in the real world.

The distributions of $(v_x,a_x)$, $(v_x,a_y)$, $(v_y,a_x)$, and $(v_y,a_y)$ for each pattern are detailed in the supplementary materials (see Appendix (a)). Besides, the patterns transition, the renderings of the real-time velocity fields, and the bird's-eye-view scenarios are all provided in our video demos of $32$ cases to help with a better intuition (see Appendix (b)).

\subsection{Transitions of Interaction Patterns}
The above section discussed the extracted interaction patterns during lane change in a spatial view. In what follows, we will gain an insight into how one interaction pattern switches to another one over time and reveal their causal relationships. Investigating lane-change behavior is usually over several independent stages of the ego vehicle's maneuvers, such as preparation, execution, and termination, but without considering interactions with other vehicles. To uncover how our learned interaction patterns behave in the stages defined by state-of-the-art approaches, we delineate the lane-change trajectory into mainly three stages as in \cite{yao2012learning}: pre-region, lane-change region, and post-region. Our focus in this paper is mainly on the transition of interaction patterns in the lane-change region, as marked in Fig. \ref{demo_dynamic}. 

Figure \ref{fig_transistion} displays the transition frequency among interaction patterns in different road contextual settings (four-lane and six-lane divided highways and their mixtures) with/without counting the self-transition frequency of patterns. Yellow and blue represent the high and low frequency of transiting between patterns, respectively. Results indicate that these interaction patterns during the lane-change region distinctly have a higher transition frequency within themselves than that between them. Corresponding analysis results of the other two regions (pre-region and post-region) are reported in the supplementary materials (see Appendix (c)).

Figure \ref{demo_dynamic} discloses the transition probability of interaction patterns straightforwardly over colored trajectories. For example, the ego vehicle in Fig. \ref{demo_dynamic}(a) is cutting into the right lane, thus giving way to a faster following vehicle on the original lane. The interaction process is decomposed into five interaction phases associated with five patterns, noted as a pattern transition chain: \#$2\rightarrow$ \#$5\rightarrow$ \#$6 \rightarrow$ \#$1\rightarrow$ \#$7$. More detailed semantic interpretations of each pattern are given in Section V-{\it D}. Correspondingly, Fig. \ref{fig_transistion}(d) presents this pattern transition chain in the transition matrix. The white dashed arrows represent the transition of interaction patterns in the pre-region and post-region, and the solid red arrows represent the transition of interaction patterns in the lane-change region. Noted that this transition matrix in Fig. \ref{fig_transistion} solely represents the transition frequencies of interaction patterns in the lane-change region, and a part of pattern \#$2$ also occurs in the lane-change region. More dynamic scenarios refer to our supplementary materials, and the associated transition chains of interaction patterns can also be shown in a transition matrix.

The whole training data in Fig. \ref{fig_transistion}(a) indicates that the most frequent self-transition pattern is of pattern \#$5$. The most frequent transition between patterns is from pattern \#$2$ to pattern \#$5$ -- the beginning of changing lanes from left to right, see Fig. \ref{demo_dynamic}(a). The second rank transition frequency is from pattern \#$4$ to pattern \#$2$, in which the ego vehicle changes to the left lane and then overtakes the surrounding vehicle, as illustrated in Fig. \ref{fig_transistion}(c). 

In the real-world traffic, it is intuitive for human drivers that changing lanes in a six-lane divided highway scenario is more challenging than in a four-lane divided highway scenario due to more uncertainties in interacting with surrounding vehicles. Fig. \ref{fig_transistion} compares the transition frequency of interaction patterns in different road contextual settings: four-lane highways in Fig. \ref{fig_transistion}(b) and (e), and six-lane highways in Fig. \ref{fig_transistion}(c) and (f). Comparisons explicitly tell that the transition matrix in the six-lane divided highway scenario contains more transition patterns than in the four-lane highway scenario. One typical transition is from pattern \#$4$ to pattern \#$2$, as shown in Fig. \ref{fig_transistion}(f).

\begin{figure}[t]
\centering
\includegraphics[width = \linewidth]{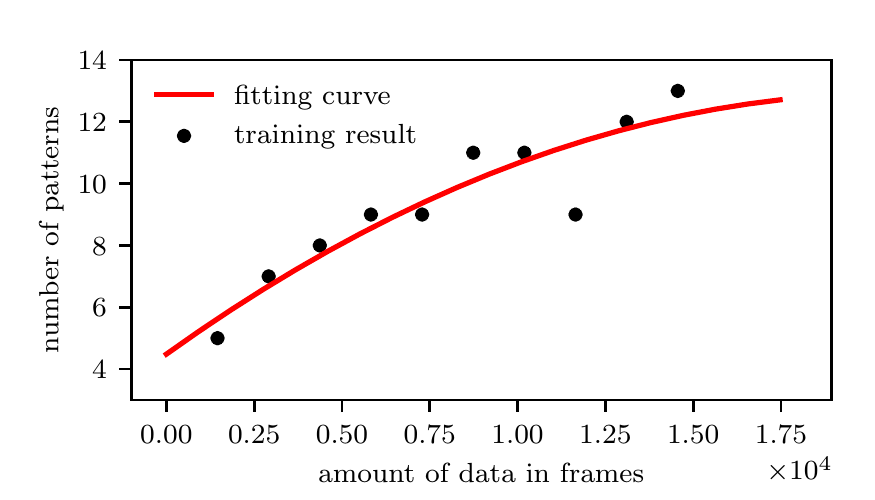}
\caption{The obtained number of interaction patterns with different amount of training data.}
\label{fig:numberofpatterns}
\end{figure}

\subsection{Further Discussion}

\subsubsection{Influence of the dataset size}
The number of interaction patterns finally learned depends on various factors, such as the training data size. As we claimed in Section I, the number of interaction patterns, theoretically, should be incrementally increasing as more new scenarios get encountered. This is the critical reason why we implemented the nonparametric model to mimic the increasing number of interactions. Fig. \ref{fig:numberofpatterns} displays the trend in the number of learned interaction patterns when increasing the training data size, which provides well-matched experimental support for the above claim.

\subsubsection{ROI definition} 
This paper predefined ROI as a rectangle area and only considered the influence of vehicles in the ROI. A large (small) area of ROI can take more (less) surrounding vehicles into account. For instance, increasing the distance front or behind the ego vehicle may change the relative velocity field since the vehicles far away from the ego vehicle may fall into the environment and impact the velocity field in the ROI. However, our proposed approach is flexible to adapt to other sizes and shapes of ROI such as polygon \cite{schmidt2019interaction} because the Gaussian process, theoretically, can describe distributions over infinite functional space in a specific spatial domain.

\subsubsection{Potential applications}
Our proposed approach decomposes a long and high-dimensional sequential interaction data into several small interpretable units (i.e., interaction patterns), as displayed in Fig. \ref{demo_dynamic}. To some extent, it compresses the complex tasks into efficient, low-dimensional representations that simplify these tasks. Besides, the Bayesian nonparametric models implemented in this paper for clustering are a general principle of representation learning \cite{niv2019learning}, i.e., clustering of experience according to the similarity to delineate task states or patterns. Although incorporating such interaction states or patterns into path/trajectory-planning and decision-making algorithms is beyond the focus of this paper, recent techniques have shown how to embed these states or patterns to improve path planning, make  safer decisions, and benefit learning efficiency of algorithms \cite{niv2019learning,galceran2017multipolicy,doshi2012bayesian}. Further, human driving behaviors cannot be fully understood without studying
the contexts in which behaviors occur\cite{rahwan2019machine}; therefore, giving an insight into interactions during lane change can provide detailed prior knowledge with autonomous vehicles to make decisions.
\section{Conclusion}\label{sec6}
This paper described a nonparametric approach that leverages the continuous and discrete stochastic processes to investigate human drivers' underlying interaction with their surrounding vehicles, for example, during a lane change on highways. To this end, we first developed a Gaussian velocity field to capture the representation of multi-vehicle interactions over the spatial space while considering the intents of surrounding vehicles. The dimension of the proposed field is insensitive to the number of vehicles in the environment. Besides, we implemented a discrete Bayesian nonparametric model that leverages a hierarchical Dirichlet process (HDP) and hidden Markov models (HMM) to learn the interaction patterns over the temporal space. We evaluated the proposed approach based on real-world data with two different traffic contextual settings.  Results showed that the proposed nonparametric approach can semantically learn the interaction patterns over the spatiotemporal space for lane change behaviors. Moreover, the proposed approach also reveals the probabilistic transitions among interaction patterns and an explicit understanding of interaction behaviors during lane change, enabling developing more efficient and safe decision-making policies of lane-change maneuvers for future studies.

\section*{Appendix}
The supplementary materials for this paper are provided via the project website: \texttt{\url{https://chengyuan-zhang.github.io/Multivehicle-Interaction}}.

(a) The distributions of $(v_x,a_x)$, $(v_x,a_y)$, $(v_y,a_x)$, and $(v_y,a_y)$ for all extracted patterns are referred to as \texttt{\href{https://chengyuan-zhang.github.io/Multivehicle-Interaction/files/Spatiotemporal_Appendix.pdf}{Spatiotemporal\_Appendix.pdf}}.

(b) The pattern transition, renderings of the real-time velocity field, and the bird's-eye-view scenarios of $32$ cases are referred to as \texttt{\url{https://youtu.be/z_vf9UHtdAM}}.

(c) The analysis of the transition matrix of the other two regions (pre-region and post-region) is referred to as the \texttt{\href{https://chengyuan-zhang.github.io/Multivehicle-Interaction/files/Spatiotemporal_Appendix.pdf}{Spatiotemporal\_Appendix.pdf}}.


%


\ifCLASSOPTIONcaptionsoff
  \newpage
\fi



{\normalem
\bibliographystyle{IEEEtran}
\bibliography{refs}}

\begin{thebibliography}{10}
\providecommand{\url}[1]{#1}
\csname url@rmstyle\endcsname
\providecommand{\newblock}{\relax}
\providecommand{\bibinfo}[2]{#2}
\providecommand\BIBentrySTDinterwordspacing{\spaceskip=0pt\relax}
\providecommand\BIBentryALTinterwordstretchfactor{4}
\providecommand\BIBentryALTinterwordspacing{\spaceskip=\fontdimen2\font plus
\BIBentryALTinterwordstretchfactor\fontdimen3\font minus
  \fontdimen4\font\relax}
\providecommand\BIBforeignlanguage[2]{{%
\expandafter\ifx\csname l@#1\endcsname\relax
\typeout{** WARNING: IEEEtran.bst: No hyphenation pattern has been}%
\typeout{** loaded for the language `#1'. Using the pattern for}%
\typeout{** the default language instead.}%
\else
\language=\csname l@#1\endcsname
\fi
#2}}

\bibitem{galceran2017multipolicy}
E.~Galceran, A.~G. Cunningham, R.~M. Eustice, and E.~Olson, ``Multipolicy
  decision-making for autonomous driving via changepoint-based behavior
  prediction: Theory and experiment,'' \emph{Autonomous Robots}, vol.~41,
  no.~6, pp. 1367--1382, 2017.

\bibitem{zhao2019influence}
C.~Zhao, W.~Wang, S.~Li, and J.~Gong, ``Influence of cut-in maneuvers for an
  autonomous car on surrounding drivers: Experiment and analysis,'' \emph{IEEE
  Transactions on Intelligent Transportation Systems},
  DOI:10.1109/TITS.2019.2914795. 2019.

\bibitem{appenzeller2017scientist}
T.~Appenzeller, ``The scientist' apprentice,'' \emph{Science}, vol. 357, no.
  6346, pp. 16--17, 2017.

\bibitem{zhang2018virtual}
J.~Zhang and A.~El~Kamel, ``Virtual traffic simulation with neural network
  learned mobility model,'' \emph{Advances in Engineering Software}, vol. 115,
  pp. 103--111, 2018.

\bibitem{leonhardt2017feature}
V.~Leonhardt and G.~Wanielik, ``Feature evaluation for lane change prediction
  based on driving situation and driver behavior,'' in \emph{2017 20th
  International Conference on Information Fusion (Fusion)}.\hskip 1em plus
  0.5em minus 0.4em\relax IEEE, 2017, pp. 1--7.

\bibitem{schwarting2019social}
W.~Schwarting, A.~Pierson, J.~Alonso-Mora, S.~Karaman, and D.~Rus, ``Social
  behavior for autonomous vehicles,'' \emph{Proceedings of the National Academy
  of Sciences}, vol. 116, no.~50, pp. 24\,972--24\,978, 2019.

\bibitem{kasper2012object}
D.~Kasper, G.~Weidl, T.~Dang, G.~Breuel, A.~Tamke, A.~Wedel, and W.~Rosenstiel,
  ``Object-oriented bayesian networks for detection of lane change maneuvers,''
  \emph{IEEE Intelligent Transportation Systems Magazine}, vol.~4, no.~3, pp.
  19--31, 2012.

\bibitem{li2018estimating}
X.~Li, W.~Wang, and M.~Roetting, ``Estimating driver's lane-change intent
  considering driving style and contextual traffic,'' \emph{IEEE Transactions
  on Intelligent Transportation Systems}, vol.~20, no.~9, pp. 3258 -- 3271,
  2018.

\bibitem{kamal2015efficient}
M.~A.~S. Kamal, S.~Taguchi, and T.~Yoshimura, ``Efficient vehicle driving on
  multi-lane roads using model predictive control under a connected vehicle
  environment,'' in \emph{2015 IEEE Intelligent Vehicles Symposium (IV)}.\hskip
  1em plus 0.5em minus 0.4em\relax IEEE, 2015, pp. 736--741.

\bibitem{do2017human}
Q.~H. Do, H.~Tehrani, S.~Mita, M.~Egawa, K.~Muto, and K.~Yoneda, ``Human
  drivers based active-passive model for automated lane change,'' \emph{IEEE
  Intelligent Transportation Systems Magazine}, vol.~9, no.~1, pp. 42--56,
  2017.

\bibitem{woo2017lane}
H.~Woo, Y.~Ji, H.~Kono, Y.~Tamura, Y.~Kuroda, T.~Sugano, Y.~Yamamoto,
  A.~Yamashita, and H.~Asama, ``Lane-change detection based on
  vehicle-trajectory prediction,'' \emph{IEEE Robotics and Automation Letters},
  vol.~2, no.~2, pp. 1109--1116, 2017.

\bibitem{yang2018time}
S.~Yang, W.~Wang, C.~Lu, J.~Gong, and J.~Xi, ``A time-efficient approach for
  decision-making style recognition in lane-changing behavior,'' \emph{IEEE
  Transactions on Human-Machine Systems}, vol.~49, no.~6, pp. 579--588, 2019.

\bibitem{liu2015classification}
P.~Liu, A.~Kurt, K.~Redmill, and U.~Ozguner, ``Classification of highway lane
  change behavior to detect dangerous cut-in maneuvers,'' in \emph{The
  Transportation Research Board (TRB) 95th Annual Meeting}, vol.~2, 2015.

\bibitem{deo2018would}
N.~Deo, A.~Rangesh, and M.~M. Trivedi, ``How would surround vehicles move? a
  unified framework for maneuver classification and motion prediction,''
  \emph{IEEE Transactions on Intelligent Vehicles}, vol.~3, no.~2, pp.
  129--140, 2018.

\bibitem{claussmann2019review}
L.~Claussmann, M.~Revilloud, D.~Gruyer, and S.~Glaser, ``A review of motion
  planning for highway autonomous driving,'' \emph{IEEE Transactions on
  Intelligent Transportation Systems}, 2019.

\bibitem{guangyu2019dbus}
M.~Guangyu~Li, B.~Jiang, Z.~Che, X.~Shi, M.~Liu, Y.~Meng, J.~Ye, and Y.~Liu,
  ``Dbus: Human driving behavior understanding system,'' in \emph{Proceedings
  of the IEEE International Conference on Computer Vision Workshops}, 2019, pp.
  0--0.

\bibitem{habibi2019incremental}
G.~Habibi, N.~Japuria, and J.~P. How, ``Incremental learning of motion
  primitives for pedestrian trajectory prediction at intersections,''
  \emph{arXiv preprint arXiv:1911.09476}, 2019.

\bibitem{niv2019learning}
Y.~Niv, ``Learning task-state representations,'' \emph{Nature neuroscience},
  vol.~22, no.~10, pp. 1544--1553, 2019.

\bibitem{lienke2019predictive}
C.~Lienke, C.~Wissing, M.~Keller, T.~Nattermann, and T.~Bertram, ``Predictive
  driving: Fusing prediction and planning for automated highway driving,''
  \emph{IEEE Transactions on Intelligent Vehicles}, vol.~4, no.~3, pp.
  456--467, 2019.

\bibitem{li2018effects}
X.~Li, W.~Wang, Z.~Zhang, and M.~R{\"o}tting, ``Effects of feature selection on
  lane-change maneuver recognition: an analysis of naturalistic driving data,''
  \emph{Journal of Intelligent and Connected Vehicles}, vol.~1, no.~3, pp.
  85--98, 2018.

\bibitem{wang2018extracting}
W.~Wang and D.~Zhao, ``Extracting traffic primitives directly from
  naturalistically logged data for self-driving applications,'' \emph{IEEE
  Robotics and Automation Letters}, vol.~3, no.~2, pp. 1223--1229, 2018.

\bibitem{hu2018probabilistic}
Y.~Hu, W.~Zhan, and M.~Tomizuka, ``Probabilistic prediction of vehicle semantic
  intention and motion,'' in \emph{2018 IEEE Intelligent Vehicles Symposium
  (IV)}.\hskip 1em plus 0.5em minus 0.4em\relax IEEE, 2018, pp. 307--313.

\bibitem{xu2017ego}
D.~Xu, X.~He, H.~Zhao, J.~Cui, H.~Zha, F.~Guillemard, S.~Geronimi, and
  F.~Aioun, ``Ego-centric traffic behavior understanding through multi-level
  vehicle trajectory analysis,'' in \emph{2017 IEEE International Conference on
  Robotics and Automation (ICRA)}.\hskip 1em plus 0.5em minus 0.4em\relax IEEE,
  2017, pp. 211--218.

\bibitem{wolf2008artificial}
M.~T. Wolf and J.~W. Burdick, ``Artificial potential functions for highway
  driving with collision avoidance,'' in \emph{2008 IEEE International
  Conference on Robotics and Automation}.\hskip 1em plus 0.5em minus
  0.4em\relax IEEE, 2008, pp. 3731--3736.

\bibitem{wang2019crash}
H.~Wang, Y.~Huang, A.~Khajepour, Y.~Zhang, Y.~Rasekhipour, and D.~Cao, ``Crash
  mitigation in motion planning for autonomous vehicles,'' \emph{IEEE
  Transactions on Intelligent Transportation Systems}, vol.~20, no.~9, pp.
  3313--3323, 2019.

\bibitem{rasekhipour2016potential}
Y.~Rasekhipour, A.~Khajepour, S.-K. Chen, and B.~Litkouhi, ``A potential
  field-based model predictive path-planning controller for autonomous road
  vehicles,'' \emph{IEEE Transactions on Intelligent Transportation Systems},
  vol.~18, no.~5, pp. 1255--1267, 2016.

\bibitem{woo2016dynamic}
H.~Woo, Y.~Ji, H.~Kono, Y.~Tamura, Y.~Kuroda, T.~Sugano, Y.~Yamamoto,
  A.~Yamashita, and H.~Asama, ``Dynamic potential-model-based feature for lane
  change prediction,'' in \emph{2016 IEEE International Conference on Systems,
  Man, and Cybernetics (SMC)}.\hskip 1em plus 0.5em minus 0.4em\relax IEEE,
  2016, pp. 000\,838--000\,843.

\bibitem{wang2016drivingssafety}
J.~Wang, J.~Wu, X.~Zheng, D.~Ni, and K.~Li, ``Driving safety field theory
  modeling and its application in pre-collision warning system,''
  \emph{Transportation research part C: emerging technologies}, vol.~72, pp.
  306--324, 2016.

\bibitem{wang2019learning}
W.~Wang, J.~Xi, and J.~K. Hedrick, ``A learning-based personalized driver model
  using bounded generalized gaussian mixture models,'' \emph{IEEE Transactions
  on Vehicular Technology}, vol.~68, no.~12, pp. 11\,679--11\,690, 2019.

\bibitem{pelleg2000x}
D.~Pelleg, A.~W. Moore, \emph{et~al.}, ``X-means: Extending k-means with
  efficient estimation of the number of clusters.'' in \emph{Icml}, vol.~1,
  2000, pp. 727--734.

\bibitem{birrell2014analysis}
S.~Birrell, J.~Taylor, A.~McGordon, J.~Son, and P.~Jennings, ``Analysis of
  three independent real-world driving studies: A data driven and expert
  analysis approach to determining parameters affecting fuel economy,''
  \emph{Transportation research part D: transport and environment}, vol.~33,
  pp. 74--86, 2014.

\bibitem{rasmussen2000infinite}
C.~E. Rasmussen, ``The infinite gaussian mixture model,'' in \emph{Advances in
  neural information processing systems}, 2000, pp. 554--560.

\bibitem{gorur2010dirichlet}
D.~G{\"o}r{\"u}r and C.~E. Rasmussen, ``Dirichlet process gaussian mixture
  models: Choice of the base distribution,'' \emph{Journal of Computer Science
  and Technology}, vol.~25, no.~4, pp. 653--664, 2010.

\bibitem{fox2007sticky}
E.~B. Fox, E.~B. Sudderth, M.~I. Jordan, and A.~S. Willsky, ``The sticky
  hdp-hmm: Bayesian nonparametric hidden markov models with persistent
  states,'' \emph{Arxiv preprint}, 2007.

\bibitem{raman2016activity}
N.~Raman and S.~J. Maybank, ``Activity recognition using a supervised
  non-parametric hierarchical hmm,'' \emph{Neurocomputing}, vol. 199, pp.
  163--177, 2016.

\bibitem{jochmann2015modeling}
M.~Jochmann, ``Modeling us inflation dynamics: A bayesian nonparametric
  approach,'' \emph{Econometric Reviews}, vol.~34, no.~5, pp. 537--558, 2015.

\bibitem{hamada2016modeling}
R.~Hamada, T.~Kubo, K.~Ikeda, Z.~Zhang, T.~Shibata, T.~Bando, K.~Hitomi, and
  M.~Egawa, ``Modeling and prediction of driving behaviors using a
  nonparametric bayesian method with ar models,'' \emph{IEEE Transactions on
  Intelligent Vehicles}, vol.~1, no.~2, pp. 131--138, 2016.

\bibitem{wang2018driving}
W.~Wang, J.~Xi, and D.~Zhao, ``Driving style analysis using primitive driving
  patterns with bayesian nonparametric approaches,'' \emph{IEEE Transactions on
  Intelligent Transportation Systems}, vol.~20, no.~8, pp. 2986--2998, 2018.

\bibitem{mahjoub2018stochastic}
H.~N. Mahjoub, B.~Toghi, and Y.~P. Fallah, ``A stochastic hybrid framework for
  driver behavior modeling based on hierarchical dirichlet process,'' in
  \emph{2018 IEEE 88th Vehicular Technology Conference (VTC-Fall)}.\hskip 1em
  plus 0.5em minus 0.4em\relax IEEE, 2018, pp. 1--5.

\bibitem{bufacchi2018action}
R.~J. Bufacchi and G.~D. Iannetti, ``An action field theory of peripersonal
  space,'' \emph{Trends in cognitive sciences}, 2018.

\bibitem{zhang2019general}
C.~Zhang, J.~Zhu, W.~Wang, and D.~Zhao, ``A general framework of learning
  multi-vehicle interaction patterns from videos,'' in \emph{2019 22nd
  International Conference on Intelligent Transportation Systems (ITSC)}.\hskip
  1em plus 0.5em minus 0.4em\relax IEEE, 2019.

\bibitem{zhu2019probabilistic}
J.~Zhu, S.~Qin, W.~Wang, and D.~Zhao, ``Probabilistic trajectory prediction for
  autonomous vehicles with attentive recurrent neural process,'' \emph{arXiv
  preprint arXiv:1910.08102}, 2019.

\bibitem{qin2019recurrent}
S.~Qin, J.~Zhu, J.~Qin, W.~Wang, and D.~Zhao, ``Recurrent attentive neural
  process for sequential data,'' \emph{arXiv preprint arXiv:1910.09323}, 2019.

\bibitem{rasmussen2003gaussian}
C.~E. Rasmussen, ``Gaussian processes in machine learning,'' in \emph{Summer
  School on Machine Learning}.\hskip 1em plus 0.5em minus 0.4em\relax Springer,
  2003, pp. 63--71.

\bibitem{bishop2006pattern}
C.~M. Bishop, \emph{Pattern recognition and machine learning}.\hskip 1em plus
  0.5em minus 0.4em\relax springer, 2006.

\bibitem{moridpour2015impact}
S.~Moridpour, E.~Mazloumi, and M.~Mesbah, ``Impact of heavy vehicles on
  surrounding traffic characteristics,'' \emph{Journal of advanced
  transportation}, vol.~49, no.~4, pp. 535--552, 2015.

\bibitem{ng2011sparse}
A.~Ng \emph{et~al.}, ``Sparse autoencoder,'' \emph{CS294A Lecture notes},
  vol.~72, no. 2011, pp. 1--19, 2011.

\bibitem{zhang2019learning}
W.~Zhang and W.~Wang, ``Learning v2v interactive driving patterns at signalized
  intersections,'' \emph{Transportation Research Part C: Emerging
  Technologies}, vol. 108, pp. 151--166, 2019.

\bibitem{taatgen2013nature}
N.~A. Taatgen, ``The nature and transfer of cognitive skills.''
  \emph{Psychological review}, vol. 120, no.~3, p. 439, 2013.

\bibitem{taniguchi2015sequence}
T.~Taniguchi, S.~Nagasaka, K.~Hitomi, N.~P. Chandrasiri, T.~Bando, and
  K.~Takenaka, ``Sequence prediction of driving behavior using double
  articulation analyzer,'' \emph{IEEE Transactions on Systems, Man, and
  Cybernetics: Systems}, vol.~46, no.~9, pp. 1300--1313, 2015.

\bibitem{fox2011sticky}
E.~B. Fox, E.~B. Sudderth, M.~I. Jordan, A.~S. Willsky, \emph{et~al.}, ``A
  sticky hdp-hmm with application to speaker diarization,'' \emph{The Annals of
  Applied Statistics}, vol.~5, no.~2A, pp. 1020--1056, 2011.

\bibitem{teh2005sharing}
Y.~W. Teh, M.~I. Jordan, M.~J. Beal, and D.~M. Blei, ``Sharing clusters among
  related groups: Hierarchical dirichlet processes,'' in \emph{Advances in
  neural information processing systems}, 2005, pp. 1385--1392.

\bibitem{johnson2013hdphsmm}
M.~J. Johnson and A.~S. Willsky, ``Bayesian nonparametric hidden semi-markov
  models,'' \emph{Journal of Machine Learning Research}, vol.~14, pp. 673--701,
  February 2013.

\bibitem{highDdataset}
R.~Krajewski, J.~Bock, L.~Kloeker, and L.~Eckstein, ``The highd dataset: A
  drone dataset of naturalistic vehicle trajectories on german highways for
  validation of highly automated driving systems,'' in \emph{2018 IEEE 21st
  International Conference on Intelligent Transportation Systems (ITSC)}, 2018.

\bibitem{yan2015classifying}
F.~Yan, L.~Weber, and A.~Luedtke, ``Classifying driver's uncertainty about the
  distance gap at lane changing for developing trustworthy assistance
  systems,'' in \emph{2015 IEEE Intelligent Vehicles Symposium (IV)}.\hskip 1em
  plus 0.5em minus 0.4em\relax IEEE, 2015, pp. 1276--1281.

\bibitem{dozat2016incorporating}
T.~Dozat, ``Incorporating nesterov momentum into adam,'' 2016.

\bibitem{yao2012learning}
W.~Yao, H.~Zhao, F.~Davoine, and H.~Zha, ``Learning lane change trajectories
  from on-road driving data,'' in \emph{2012 IEEE Intelligent Vehicles
  Symposium}.\hskip 1em plus 0.5em minus 0.4em\relax IEEE, 2012, pp. 885--890.

\bibitem{schmidt2019interaction}
M.~Schmidt, C.~Manna, J.~H. Braun, C.~Wissing, M.~Mohamed, and T.~Bertram, ``An
  interaction-aware lane change behavior planner for automated vehicles on
  highways based on polygon clipping,'' \emph{IEEE Robotics and Automation
  Letters}, vol.~4, no.~2, pp. 1876--1883, 2019.

\bibitem{doshi2012bayesian}
F.~Doshi-Velez, ``Bayesian nonparametric approaches for reinforcement learning
  in partially observable domains,'' Ph.D. dissertation, Massachusetts
  Institute of Technology, 2012.

\bibitem{rahwan2019machine}
I.~Rahwan, M.~Cebrian, N.~Obradovich, J.~Bongard, J.-F. Bonnefon, C.~Breazeal,
  J.~W. Crandall, N.~A. Christakis, I.~D. Couzin, M.~O. Jackson, \emph{et~al.},
  ``Machine behaviour,'' \emph{Nature}, vol. 568, no. 7753, pp. 477--486, 2019.

\end{thebibliography}
%
%

%

\begin{IEEEbiography}[{\includegraphics[width=1in,height=1.25in,clip,keepaspectratio]{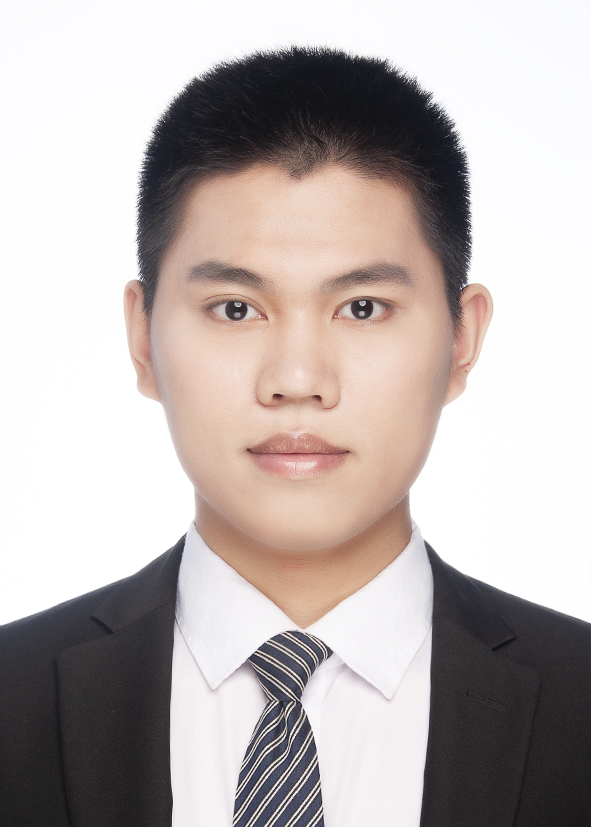}}]{Chengyuan Zhang} is a graduate student with the Department of Civil Engineering, McGill University. He received his B.S. degree in Automotive Engineering from Chongqing University, Chongqing, China, in 2019. His research interests are Bayesian learning, macro/micro driving behavior analysis, intelligent transportation systems, and autonomous vehicles.
\end{IEEEbiography}

\begin{IEEEbiography}[{\includegraphics[width=1in,height=1.25in,clip,keepaspectratio]{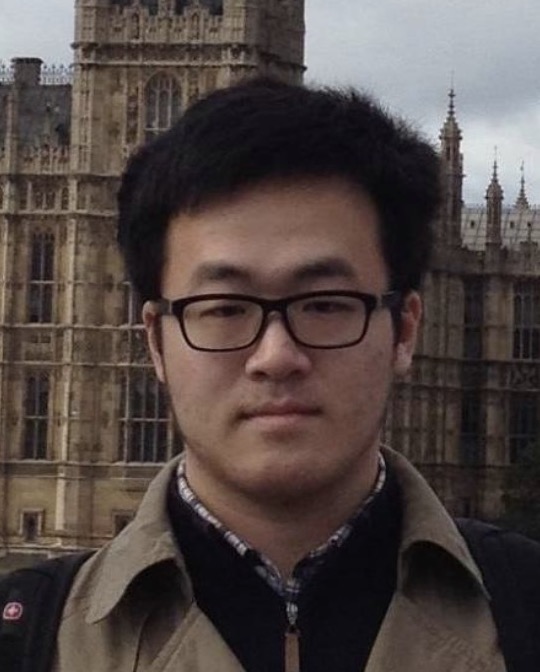}}]{Jiacheng Zhu}
is a Ph.D student with the Department of Mechanical Engineering, Carnegie Mellon University. 
He is interested in building machine learning system that can operate with limited prior knowledge and can adapt to unobserved circumstances. He approaches this goal using probabilistic modeling and approximate inferences. Some of his research topics include (deep) probabilistic graphical models, varaitional inferences, and traffic scenario recognition. 
\end{IEEEbiography}

\begin{IEEEbiography}[{\includegraphics[width=1in,height=1.25in,clip,keepaspectratio]{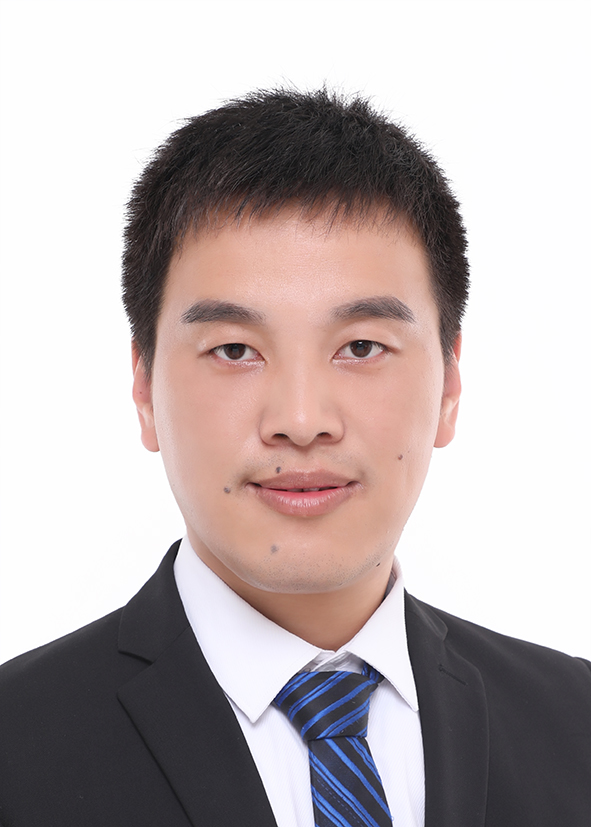}}]{Wenshuo Wang} (S'15-M'18) received his Ph.D. degree in Mechanical Engineering from Beijing Institute of Technology, Beijing, China in 2018. Currently, he is a Postdoctoral Researcher at the California Partners for Advanced Transportation Technology (PATH), University of California at Berkeley (UC Berkeley). Before joint UC Berkeley, he has worked as a Postdoctoral Associate at Carnegie Mellon University (CMU) from 2018 -- 2019. He also worked as a Research Scholar with UC Berkeley from 2015 -- 2017 and with the University of Michigan, Ann Arbor, from 2017 -- 2018. His current research focus is on Bayesian nonparametric learning, reinforcement learning on multi-agent interaction behavior modeling and prediction at human-levels for autonomous vehicles, intelligent transportation systems in common-but-challenging situations.
\end{IEEEbiography}

\begin{IEEEbiography}[{\includegraphics[width=1in,height=1.25in,clip,keepaspectratio]{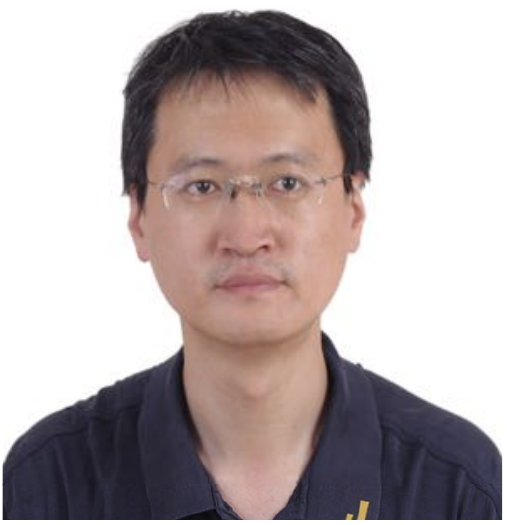}}]{Junqiang Xi} received the B.S. in Automotive Engineering from Harbin Institute of Technology, Harbin, China, in 1995 and the PhD in Vehicle Engineering from Beijing Institute of Technology (BIT), Beijing, China, in 2001. In 2001, he joined the State Key Laboratory of Vehicle Transmission, BIT. During 2012-2013, he made research as an advanced research scholar in Vehicle Dynamic and Control Laboratory, Ohio State University(OSU), USA. He is Professor and Director of Automotive Research Center in BIT currently. His research interests include vehicle dynamic and control, power-train control, mechanics, intelligent transportation system and intelligent vehicles.
\end{IEEEbiography}







\end{document}